\documentclass[12pt,twoside]{report}

\newcommand{\reporttitle}{LLMs to Support a Domain Specific Knowledge Assistant}
\newcommand{\reportauthor}{Maria-Flavia Lovin}
\newcommand{\supervisor}{Dr. Pedro Baiz Villafranca}
\newcommand{\secondmarker}{Dr. Tolga Birdal}
\newcommand{\degreetype}{MSc Artificial Intelligence}

\usepackage[a4paper,hmargin=2.8cm,vmargin=2.0cm,includeheadfoot]{geometry}
\usepackage{textpos}
\usepackage[numbers]{natbib} 
\usepackage{tabularx,longtable,multirow,subfigure,caption}
\usepackage{fncylab} 
\usepackage{fancyhdr} 
\usepackage{url} 
\usepackage[english]{babel}
\usepackage{amsmath}
\usepackage{graphicx}
\usepackage{dsfont}
\usepackage{epstopdf} 
 \usepackage{array}
\usepackage{latexsym}
\usepackage{appendix}
\usepackage[hypertexnames=false,colorlinks]{hyperref}
\usepackage{listings}
\usepackage{amsmath}
\usepackage{amssymb}

\hypersetup{pdftitle={},
  pdfsubject={}, 
  pdfauthor={},
  pdfkeywords={}, 
  pdfstartview=FitH,
  pdfpagemode={UseOutlines},
  bookmarksnumbered=true, bookmarksopen=true, colorlinks,
    citecolor=black,%
    filecolor=black,%
    linkcolor=black,%
    urlcolor=black}

\usepackage[all]{hypcap}

\renewcommand{\sectionmark}[1]{}
\def\@makechapterhead#1{%
  \vspace*{10\p@}%
  {\parindent \z@ \raggedright \sffamily
    \interlinepenalty\@M
    \Huge\bfseries \thechapter \space\space #1\par\nobreak
    \vskip 30\p@
  }}

\def\@makeschapterhead#1{%
  \vspace*{10\p@}%
  {\parindent \z@ \raggedright
    \sffamily
    \interlinepenalty\@M
    \Huge \bfseries  #1\par\nobreak
    \vskip 30\p@
  }}

\allowdisplaybreaks

\usepackage{algorithm}
\usepackage{algpseudocode}
\usepackage{listings} 
\lstdefinelanguage{json}{
    basicstyle=\normalfont\ttfamily\footnotesize,
    numbers=left,
    numberstyle=\scriptsize,
    stepnumber=1,
    numbersep=8pt,
    showstringspaces=false,
    breaklines=true,
    frame=single,
    backgroundcolor=\color{background},
    literate=
     *{0}{{{\color{numb}0}}}{1}
      {1}{{{\color{numb}1}}}{1}
      {2}{{{\color{numb}2}}}{1}
      {3}{{{\color{numb}3}}}{1}
      {4}{{{\color{numb}4}}}{1}
      {5}{{{\color{numb}5}}}{1}
      {6}{{{\color{numb}6}}}{1}
      {7}{{{\color{numb}7}}}{1}
      {8}{{{\color{numb}8}}}{1}
      {9}{{{\color{numb}9}}}{1}
      {:}{{{\color{punct}{:}}}}{1}
      {,}{{{\color{punct}{,}}}}{1}
      {\{}{{{\color{delim}{\{}}}}{1}
      {\}}{{{\color{delim}{\}}}}}{1}
      {[}{{{\color{delim}{[}}}}{1}
      {]}{{{\color{delim}{]}}}}{1},
}
\usepackage{xcolor}        
\colorlet{punct}{red!60!black}
\definecolor{background}{HTML}{EEEEEE}
\definecolor{delim}{RGB}{20,105,176}
\colorlet{numb}{magenta!60!black}

\date{September 2024}

\begin{document}

\begin{titlepage}

\newcommand{\HRule}{\rule{\linewidth}{0.5mm}}

\includegraphics[width = 4cm]{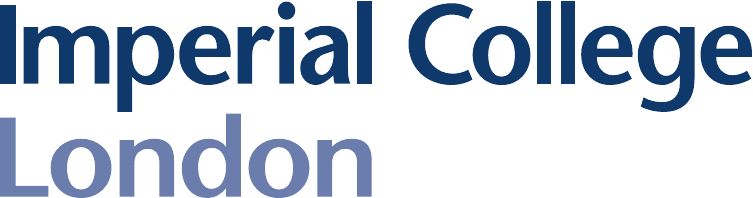}\\[0.5cm] 

\center 

\textsc{\Large Imperial College London}\\[0.5cm] 
\textsc{\large Department of Computing}\\[0.5cm] 

\HRule \\[0.4cm]
{ \huge \bfseries \reporttitle}\\ 
\HRule \\[1.5cm]

\begin{minipage}{0.4\textwidth}
\begin{flushleft} \large
\emph{Author:}\\
\reportauthor 
\end{flushleft}
\end{minipage}
~
\begin{minipage}{0.4\textwidth}
\begin{flushright} \large
\emph{Supervisor:} \\
\supervisor \\ 
\emph{Second marker:}\\
\secondmarker
\end{flushright}
\end{minipage}\\[4cm]

\vfill
Submitted in partial fulfillment of the requirements for the MSc degree in
\degreetype~of Imperial College London\\[0.5cm]

\makeatletter
\@date 
\makeatother

\end{titlepage}

\pagenumbering{roman}
\setcounter{page}{1}
\pagestyle{fancy}

\begin{abstract}
This work presents a custom approach to developing a domain-specific knowledge assistant for sustainability reporting using the International Financial Reporting Standards (IFRS). In this domain, there is no publicly available question-answer dataset, which has impeded the development of a high-quality chatbot to support companies with IFRS reporting. The two key contributions of this project therefore are:\\

(1) A high-quality synthetic question-answer dataset based on IFRS sustainability standards, created using a novel generation and evaluation pipeline leveraging Large Language Models (LLMs). This dataset comprises 1,063 diverse question-answer pairs that address a wide spectrum of potential user queries in sustainability reporting. Various LLM-based techniques are employed to create the dataset, including chain-of-thought reasoning and few-shot prompting. A custom evaluation framework is developed to assess question and answer quality across multiple dimensions, including faithfulness, relevance, and domain specificity. The resulting dataset averages a score range of 8.16 out of 10 on these metrics.\\

(2) Two custom architectures tailored for question-answering in the sustainability reporting domain - a retrieval augmented generation (RAG) pipeline and a fully LLM-based pipeline. The architectures are developed by experimenting, fine-tuning, and training on the question-answer dataset. The final pipelines feature an LLM fine-tuned on domain-specific data and an industry classification component to improve the handling of complex, multi-industry queries. The RAG architecture achieves an accuracy of 85.32\% on single-industry and 72.15\% on cross-industry multiple-choice questions, outperforming the baseline approach by 4.67 and 19.21 percentage points, respectively. The LLM-based pipeline achieves an accuracy of 93.45\% on single-industry and 80.30\% on cross-industry multiple-choice questions, an improvement of 12.80 and 27.36 percentage points over the baseline, respectively.  \\

\end{abstract}

\section*{Acknowledgments}

I extend my deepest appreciation to Dr. Pedro Baiz Villafranca for his invaluable guidance and support throughout this project. His encouragement and constructive feedback from the project's inception have been instrumental in its success.\\

I also wish to thank Dr. Tolga Birdal for his co-supervision and for providing clarity on a number of questions during the early stages of this project.\\

Lastly, I want to express my gratitude to my family and close friends for their continuous love and support. Without them, this work would not have been possible.

\tableofcontents 

\newpage

\pagenumbering{arabic}
\setcounter{page}{1}

\chapter{Introduction}

\section{Motivation}\label{motivations}

In recent years, businesses worldwide have undergone a shift toward sustainable practices, driven by increasing environmental concerns, regulatory pressures, and stakeholder expectations \cite{ey}. In response to this, the International Financial Reporting Standards (IFRS) Foundation introduced sustainability-related disclosure standards that provide a comprehensive framework for companies to report on their environmental, social, and governance (ESG) impacts \citep{ifrsorgc9:online}. However, the complexity and evolving nature of these standards present challenges for organizations that strive to produce compliant sustainability reports. In particular, the preparation of such reports requires deep domain knowledge, the ability to interpret and apply complex guidelines, and an effort to keep up to date on changes to standards over time. This can present a barrier for companies that lack the resources or expertise to navigate these requirements effectively, which, in turn, can hinder compliance and impede the broader goal of promoting sustainable business practices.\\

This setting presents a much-needed application for AI to assist companies with knowledge of IFRS reporting standards. In fact, a growing interest is expressed in developing intelligent assistants for a wide range of corporate applications \cite{nicolescu2022human, kim2024ragqaragintegratinggenerative}. However, no work has been published to date developing such a system in IFRS sustainability reporting, despite this being a domain that is relevant to thousands of companies worldwide. This has resulted in a two-fold research gap: there are no published, high-quality question-answering systems based on this information, and there is no publicly available dataset that can be utilised to train and evaluate such systems in this domain. The lack of a dataset presents a hurdle towards implementing AI question-answering systems as they cannot be trained and evaluated appropriately. Simultaneously, the creation of such a dataset, whether by humans or by Large Language Models (LLMs), is a challenging task. The problem, therefore, becomes self-perpetuating.

\section{Objectives}
The aim of this project is to design, implement, and evaluate a question-answering system within the IFRS sustainability reporting domain. Specifically, we first strive to leverage LLMs to make a question-answer (QA) dataset that we can then use to develop an architecture for answering user queries on this topic. The objectives of this project are to: 

\begin{enumerate}
    \item Define the knowledge scope of the assistant and the typical queries a user would ask within this domain.
    \item Design and implement a QA generation and evaluation pipeline using state-of-the-art LLMs.
    \item Produce a synthetic QA dataset based on the IFRS sustainability reporting standards.
    \item Leverage LLMs and retrieval augmented generation (RAG) systems to develop a robust architecture for question-answering on the sustainability reporting domain.
    \item Evaluate and optimise the system's performance on a diverse range of user queries.
\end{enumerate}

\section{Contributions}

This work presents a two-fold solution to the problem defined in Section \ref{motivations}. We present a first-of-its-kind test suite for evaluating LLMs on the question-answering task within the sustainability reporting domain and design an assistant that provides accurate guidance on sustainability reporting by leveraging advancements in LLMs and RAG. This project presents several contributions:
\begin{enumerate}
    \item A novel dataset of diverse, synthetically generated question-answer (QA) pairs, in both multiple-choice and free-text format, based on IFRS Sustainability Reporting Standards.
    \item A custom QA generation and evaluation pipeline.
    \item Custom LLM-based evaluation metrics for quality control of synthetically-generated QA pairs.
    \item A systematic evaluation of language models and RAG methods using the synthesised QA dataset.
    \item Two custom architectures for sustainability reporting question-answering - a RAG-based pipeline and a fully LLM-based pipeline. Both of these integrate an industry classifier and an LLM fine-tuned on the QA dataset.
\end{enumerate}

\chapter{Background and Related Work}

\section{Advancements in Natural Language Processing}

This section explores key developments in Natural Language Processing (NLP) and Large Language Models (LLMs) that have provided the foundation for domain-specific question-answering systems.

\subsection{Large Language Models}

The introduction of the Transformer architecture \cite{vaswani2023attention} marked a significant shift in NLP, moving away from recurrent neural networks towards models based on self-attention mechanisms. These mechanisms allow for efficient parallel processing and excel at capturing long-range dependencies in text, which are crucial for understanding complex language structures. This shift has enabled the development of sophisticated pre-trained models that can be fine-tuned for various tasks, significantly improving performance across the board.

\subsection{BERT}

BERT (Bidirectional Encoder Representations from Transformers) \cite{devlin2019bertpretrainingdeepbidirectional} utilises the Transformer architecture's encoder to understand text context bidirectionally. It is pre-trained on tasks like Masked Language Modeling, where it predicts masked tokens, and Next Sentence Prediction, enhancing its ability to capture nuanced language features. BERT's architecture, with its deep bidirectional context understanding, makes it particularly effective for tasks requiring deep comprehension, such as sentiment analysis and text classification. Its design allows for straightforward adaptation to classification tasks by adding a task-specific output layer for fine-tuning. This flexibility and effectiveness have led to BERT's widespread adoption in various NLP applications, including named entity recognition and question answering.

\subsection{GPT}

The Generative Pre-trained Transformer (GPT) \cite{radford2018improving} is an auto-regressive language model that leverages the Transformer architecture's decoder. It processes text in a unidirectional manner, generating coherent text by predicting each word based on the preceding context. GPT is pre-trained on extensive datasets using unsupervised learning, optimising its parameters to capture complex language patterns. This pre-training equips GPT to perform various NLP tasks such as text completion, summarization, and more. The model's ability to generate human-like text has made it a popular choice for applications in creative writing, dialogue systems, and other generative tasks.

\subsection{Limitations of Large Language Models}

Despite their impressive capabilities, LLMs have inherent limitations. They can struggle with queries requiring current information beyond their training data and may generate inaccurate or nonsensical text, known as ``hallucinations" \cite{zhang2023siren}. Additionally, LLMs often lack true understanding and can find multi-step reasoning challenging. While larger models exhibit improved reasoning abilities, integrating genuine causal reasoning remains an active research area. The frequency and representation of specific documents or facts in their training data can significantly impact their ability to answer related questions accurately. Moreover, the computational resources required for training and deploying these models can be substantial, which can pose challenges for scalability and accessibility.

\section{Prompt Engineering}

This section provides an overview of state-of-the-art prompt engineering techniques used in LLM-based systems.

\subsection{Prompt Templates}

Large language models typically operate by generating text outputs in response to user-provided input prompts, which can be natural language instructions, questions, or other forms of content such as images or audio \cite{schulhoff2024promptreportsystematicsurvey}. Prompt engineering is a method of utilising LLMs for various tasks by crafting input prompts to guide LLMs towards desired outputs \cite{schulhoff2024promptreportsystematicsurvey}. It relies on the principle that LLMs generate responses based on patterns in their training data, so a carefully crafted prompt can activate relevant knowledge within the model's parameters. This process often involves providing context, specifying desired output formats, and framing tasks in ways that align with the model's training. Early work in prompt engineering focused on techniques such as including examples or explicit instructions within prompts \cite{brown2020language}. Later research focused on standardising prompt creation and improving reproducibility using methods such as meta-prompting \cite{reynolds2021prompt, zhou2022large} (prompting an LLM to generate or improve a prompt or prompt template) and using prompt templates \cite{shin2020autoprompt}. \\

Prompt templates are functions containing variables that can be replaced with specific content to create customised inputs - or template instances - for various tasks \cite{shin2020autoprompt}. Prompts may be composed of different components, the key one being the `directive' or `intent' - this is the instruction or question that forms the core of the prompt. Additional components include examples (also known as `shots'), output formatting (such as in markdown or JSON formats), style instructions (used to modify the style rather than the structure of the output), role (a persona given to the LLM that can improve writing style), and additional information that gives the LLM context for the text it is generating \cite{schulhoff2024promptreportsystematicsurvey}.\\

Prompt templates can be designed to incorporate further prompt engineering techniques, such as Chain-of-Thought and Few-shot learning.

\subsection{In-Context Learning (ICL)} 

In-context learning (ICL) refers to LLMs' ability to adapt to new tasks by utilising examples or instructions embedded in the input prompt, eliminating the need for model retraining or parameter adjustments \cite{brown2020language}. Two prominent ICL techniques are chain-of-thought and few-shot prompting.\\

Chain-of-Thought (CoT) prompting \cite{wei2022chain} is a technique for improving the reasoning capabilities of LLMs by prompting the model to generate intermediate steps or thoughts leading to a final answer, rather than directly producing the end result. The concept of CoT prompting was introduced to address the limitations of LLMs in tasks requiring multi-step reasoning or complex problem-solving and has been shown to improve LLM performance on tasks in domains such as mathematics, logic, and analytical reasoning. \\

A typical implementation of CoT prompting involves structuring the input prompt to include examples of step-by-step reasoning and either explicitly or implicitly encouraging the model to emulate a similar thought process in its response. Drawbacks of the approach include that factors such as model size and the complexity of the reasoning required can influence the success of this technique, while generating and validating appropriate CoT prompts can be challenging and time-consuming, particularly for specialised domains. Recent work has explored variations of CoT prompting, including zero-shot CoT \cite{kojima2023largelanguagemodelszeroshot}, which improves LLM performance by incorporating phrases such as ``Let’s think step by step".\\

CoT is a special application of few-shot prompting \cite{brown2020language}, which is a technique whereby several examples are provided within the prompt to guide the language model's response. The principle behind few-shot prompting is to provide task-specific context through examples to help the model understand the task requirements and generate more accurate responses. The effectiveness of few-shot prompting has been demonstrated across various natural language processing tasks, including text classification, question answering, and language translation, and has shown particular promise in scenarios where task-specific training data is limited or when rapid adaptation to new tasks is required. Nevertheless, designing an effective few-shot prompt is a challenging task, as several factors such as exemplar ordering and format may affect outcomes \cite{schulhoff2024promptreportsystematicsurvey}.

\subsection{Prompting for LLM-based Evaluation}

Recent research has explored the use of LLMs as evaluators due to their ability to extract and reason about information, as well as to understand the intent of the users \cite{schulhoff2024promptreportsystematicsurvey}. The prompting techniques used incorporate ICL \cite{kocmi2023gemba}, role-based evaluation \cite{wu2023large}, and CoT prompting \cite{lu2023error}. Output frameworks for evaluation have also been shown to significantly impact the performance of LLMs as evaluators \cite{gao2023human}. A particularly significant component of the output framework is the styling of the output - such as in markdown or JSON format - which has been shown to make the LLM's judgment more accurate \cite{hada2023large, lin2023llm}. Other output formats for evaluation include linear scales, binary scores, and Likert scales \cite{schulhoff2024promptreportsystematicsurvey}.

\section{LoRA Fine-Tuning}

Low-Rank Adaptation (LoRA) \cite{hu2021loralowrankadaptationlarge} is a parameter-efficient fine-tuning method for LLMs. This technique addresses the computational challenges associated with fine-tuning large pre-trained models by introducing a small number of trainable parameters while keeping the majority of the model frozen.\\

The core idea behind LoRA is to approximate the weight updates during fine-tuning using low-rank decomposition. Instead of updating the full weight matrices, LoRA introduces smaller matrices that capture the changes in a low-dimensional space. These additional matrices are then used to modify the output of the original model layers through simple matrix multiplication.\\

Formally, for a pre-trained weight matrix $W \in \mathbb{R}^{d \times k}$, LoRA defines the update as:

\[
h = Wx + BAx
\]

where $A \in \mathbb{R}^{r \times d}$ and $B \in \mathbb{R}^{d \times r}$ are the low-rank matrices, and $r$ is the chosen rank, typically much smaller than $d$ and $k$. Only $A$ and $B$ are trained during the fine-tuning process, significantly reducing the number of trainable parameters.\\

LoRA has demonstrated comparable performance to full fine-tuning on various natural language processing tasks while requiring substantially less computational resources \cite{hu2021loralowrankadaptationlarge}, making it useful for adapting LLMs to specific domains or tasks. Nonetheless, its effectiveness can vary depending on the degree of domain shift from the pre-training data, the choice of which layers to adapt, and the rank of the decomposition.

\section{Retrieval Augmented Generation}

Retrieval Augmented Generation (RAG) \cite{lewis2021retrievalaugmented} is a method for enhancing the capabilities of LLMs by grounding them in external knowledge sources. At its core, RAG operates on the principle of retrieving relevant information from a knowledge base and using this information to augment the input to a language model. It is a hybrid approach that combines the strengths of retrieval and generation tasks in NLP to address limitations in traditional language models, such as hallucination and the inability to access proprietary and/or up-to-date information. This section describes the basic RAG pipeline and explores recent advancements and challenges in this field.\\

\begin{figure}[H]
    \centering
    \includegraphics[width=\linewidth]{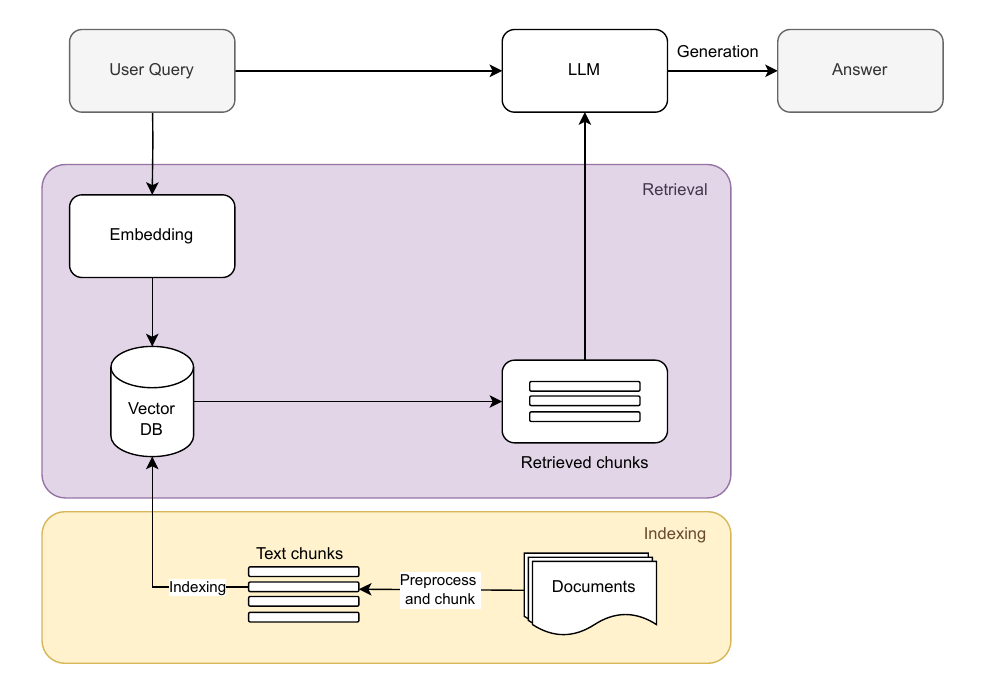}
    \caption{Basic RAG pipeline.}
    \label{fig:basic_rag}
\end{figure}

The basic RAG pipeline, shown in Figure \ref{fig:basic_rag}, consists of three main components - data indexing, retrieval, and generation.

\subsubsection{Data Indexing}

Data indexing is the first step in the RAG pipeline and involves preprocessing the documents, segmenting them, and embedding these segments to be stored in a vector database. The chunking methods and embedding models selected can have significant impacts on the performance of the RAG system.\\

Chunking defines the granularity of the retrieved context - coarse-grained segments potentially offer more comprehensive information but risk including irrelevant content, while fine-grained units may preserve semantic integrity but increase retrieval complexity. On this spectrum, many chunking strategies exist, from simple fixed-length chunks to more advanced semantic approaches, created to optimise retrieval granularity, which can significantly enhance both retrieval efficacy and downstream task performance \cite{gao2024retrievalaugmented}.\\

The second step involves embedding the chunks by converting the text information into dense vector representations. The embeddings capture semantic meaning in a compact form, allowing for rapid similarity comparisons \cite{gao2024retrievalaugmented}. Typically, the embedding process utilises pre-trained language models or specialised embedding models. For instance, embedding models trained by OpenAI \cite{openaiembeddingmodel} and Voyage AI \cite{voyageembeddingmodel} have gained popularity due to their ability to generate high-quality sentence embeddings (the latter model is used by Anthropic for their LLMs). \\

To index the data within the vector database, metadata can be appended to each chunk with relevant information about the content of the chunk, such as what source file and page number it is taken from. This allows subsequent filtering of the chunks based on this metadata to limit the scope of the retrieval \cite{gao2024retrievalaugmented}.

\subsubsection{Retrieval}

The retrieval component is responsible for identifying the most relevant information from the embedded knowledge base given a particular query or context. Traditional retrieval methods rely on embedding cosine similarity. Semantic cosine similarity \cite{rahutomo2012semantic} measures the angular proximity of non-zero vectors in a vector space, ranging from -1 to 1. It is predominantly used in text vector matching to assess both semantic and structural similarities between texts and is especially effective in high-dimensional spaces due to its focus on vector direction rather than magnitude. Beyond cosine similarity, dense retrieval methods have gained prominence in recent years. Karpukhin et al. \cite{karpukhin2020dense} introduced Dense Passage Retrieval (DPR), which uses separate encoders for queries and passages, trained end-to-end on relevant data. This approach has shown superior performance on complex queries where semantic understanding is crucial. More recent work has focused on hybrid retrieval systems that integrate keyword, semantic, and
vector searches \cite{gao2024retrievalaugmented}. \\

One common problem among these approaches is their reliance on the user's original query for retrieval, as the query may be poorly phrased or misrepresentative of the user's true intent \cite{gao2024retrievalaugmented}. To address this issue, several query optimisation techniques have been developed, including query expansion and transform methods like multi-query \cite{medium_rag_not_working}, sub-query \cite{zhou2023leasttomostpromptingenablescomplex}, and query re-writing using LLMs or smaller specialised models. Query optimisation aims to improve the relevance of the user query to the true context needed to answer it, thereby improving retrieval by similarity measures like embedding cosine similarity.

\subsubsection{Generation}

Once context is retrieved, it is passed to a pre-trained model along with the initial query to generate a response to the query based on the context. Answer generation can vary depending on the specific task, the most important consideration being whether the model utilises its parametric knowledge or is constrained to the information provided in the retrieved context \cite{gao2024retrievalaugmented}. Fine-tuning is an approach to embed knowledge into the generator LLM to improve its performance on downstream tasks such as question-answering.

\subsubsection{RAG versus Fine-tuning}

RAG and fine-tuning are two approaches to LLM optimisation. RAG utilises real-time knowledge retrieval, which is advantageous in domains where information is updated frequently and interpretability is valued. However, RAG typically incurs higher latency owing to all the steps required to retrieve context before generating a response. Fine-tuning, on the other hand, modifies model parameters, which allows deeper customization of model behaviour but requires retraining for updates and significant computational resources.\\

Empirical studies \cite{gao2024retrievalaugmented} indicate that RAG shows superior performance in knowledge-intensive tasks compared to unsupervised fine-tuning, particularly for new information, and that LLMs show limited ability to learn new facts through unsupervised fine-tuning alone. Nonetheless, the choice between RAG and fine-tuning depends on the specific requirements of the application, including data dynamics, the need for customisation, and the computational resources available. It is important to note that these methods are not mutually exclusive and can be combined to potentially enhance overall model performance.

\section{Evaluation of LLM Generated Text}

Evaluating the performance of domain-specific question-answering systems built using RAG methods presents unique challenges. This section discusses various evaluation metrics and the specific difficulties associated with assessing knowledge-intensive dialogue systems.

\subsection{NLP Evaluation Metrics}

Traditional NLP evaluation metrics have been crucial in assessing the quality of LLM-generated text across various tasks. Two of the most widely used metrics are BLEU and ROUGE.\\

BLEU (Bilingual Evaluation Understudy) \cite{papieni_bleu} is primarily used for quality evaluation in machine translation. It measures the precision of n-gram matches between the generated text and one or more reference texts. BLEU scores range from 0 to 1, with higher scores indicating better quality translations. The metric computes the geometric mean of n-gram precisions (typically up to 4-grams) and applies a brevity penalty to discourage very short translations. \\

\begin{equation}
    \text{BLEU} = BP \cdot \exp\left(\sum_{n=1}^N w_n \log p_n\right)
\end{equation}

\noindent where $BP$ is the brevity penalty, $N$ is the maximum n-gram order (typically 4), $w_n$ are positive weights summing to 1, and $p_n$ is the modified n-gram precision. The brevity penalty is calculated as:

\begin{equation}
    BP = 
    \begin{cases}
        1 & \text{if } c > r \\
        e^{(1-r/c)} & \text{if } c \leq r
    \end{cases}
\end{equation}

\noindent where $c$ is the length of the candidate translation and $r$ is the effective reference length.\\

ROUGE (Recall-Oriented Understudy for Gisting Evaluation), \cite{lin2004rouge}, is commonly used in text summarisation tasks. Several ROUGE variants exist, with ROUGE-N and ROUGE-L being the most popular. ROUGE-N measures the recall of n-grams between the generated summary and reference summaries, while ROUGE-L considers the longest common subsequence (LCS) and calculates the precision, recall, and F1-score based on the length of the LCS. ROUGE-N focuses on the grammatical correctness and fluency of text, while ROUGE-L focuses on semantic similarity. ROUGE scores typically range from 0 to 1, with higher scores indicating better summarisation quality.\\

\begin{equation}
    \text{ROUGE-N} = \frac{\sum_{S \in \{\text{Reference Summaries}\}} \sum_{\text{gram}_n \in S} \text{Count}_\text{match}(\text{gram}_n)}{\sum_{S \in \{\text{Reference Summaries}\}} \sum_{\text{gram}_n \in S} \text{Count}(\text{gram}_n)}
\end{equation}

\noindent where $\text{Count}_\text{match}(\text{gram}_n)$ is the maximum number of n-grams co-occurring in a candidate summary and a set of reference summaries, and $\text{Count}(\text{gram}_n)$ is the number of n-grams in the reference summaries.\\

\begin{equation}
    \text{ROUGE-L} = \frac{(1 + \beta^2)R_L P_L}{R_L + \beta^2 P_L}
\end{equation}
\noindent where $R_L$ is the LCS-based recall, $P_L$ is the LCS-based precision, and $\beta$ is a parameter to control the importance of recall and precision (usually $\beta = 1.2$).\\

\noindent The components $R_L$ and $P_L$ are defined as:
\begin{equation}
    R_L = \frac{\text{LCS}(X,Y)}{\text{len}(X)}
\end{equation}
\begin{equation}
    P_L = \frac{\text{LCS}(X,Y)}{\text{len}(Y)}
\end{equation}
\noindent where $\text{LCS}(X,Y)$ is the length of the Longest Common Subsequence between the reference summary $X$ and the candidate summary $Y$, and $\text{len}(X)$ and $\text{len}(Y)$ are the lengths of $X$ and $Y$ respectively.\\

These metrics have been widely used to assess the quality of generated text. However, they have limitations, particularly in the context of conversational AI and knowledge-intensive tasks. Their reliance on exact word matches or predefined synonyms can fail to capture more nuanced semantic similarities, especially in open-ended generation tasks.\\

More recent metrics like BERTScore \cite{zhang2019bertscore} and BLEURT \cite{sellam2020bleurt} leverage pre-trained language models to capture the semantic similarity between generated and reference texts, showing better correlation with human judgments compared to traditional n-gram-based metrics. BERTScore computes cosine similarities between BERT embeddings of generated and reference texts, while BLEURT fine-tunes BERT on human judgments to learn a task-specific metric. 

\subsection{RAG Evaluation}

Recent research has explored various approaches to evaluate the quality and factuality of text generated by LLMs based on a particular source context. One major focus has been on using LLMs themselves to detect hallucinations and estimate faithfulness. Zhang et al. \cite{zhang2023interpretable} suggested using few-shot prompting strategies for predicting factuality, while \cite{min2023factscore} proposed linking generated responses to external knowledge bases. LLMs have also been used to evaluate other aspects of generated text beyond factuality. Fu et al. \cite{fu2023gptscore} introduced GPTScore, which uses generative pre-trained models to score generated texts using prompting techniques such as zero-shot instruction. Wang et al. \cite{wang2023chatgpt} explored directly asking ChatGPT to provide numerical ratings for specific aspects of answers. \\

The evaluation of RAG architectures in particular is a challenging task, requiring many dimensions to be considered. These include the relevance and conciseness of selected context passages, the LLM's ability to accurately interpret and utilise these passages, and the overall quality of the generated output. Frameworks such as RAGAS \cite{es2023ragas} have been developed, proposing a suite of LLM-based evaluation metrics for RAG systems as a whole. \\

Nevertheless, LLM-based evaluation methods have limitations, such as being sensitive to factors like prompt design and presentation order \cite{wang2023large}. To address these limits, recent work by \cite{li2023halueval} has proposed a multi-dimensional evaluation framework for knowledge-intensive question-answering systems that combines automated metrics with human evaluation across dimensions such as factual accuracy, relevance, coherence, and user satisfaction.

\section{Synthetic Question-Answer Data Generation}

This section reviews the literature on the automatic generation of synthetic question-answer (QA) data. QA data is used for training and/or evaluating RAG systems that specialise in Open Domain Question Answering (ODQA) tasks. Automatic generation of synthetic QA data to improve these systems has been an active area of research for decades \cite{mitkov_mcq} \cite{rus2010first}, and is particularly useful in domains where QA datasets do not already exist and/or where curating a human (expert) annotated dataset would involve prohibitive time and monetary costs.\\

However, automatic QA generation is a highly complex task and is particularly difficult for domain-specific RAG systems, where questions must be tailored to test the quality of the system within the context of a specific use case. This is because the questions must be representative of queries that the target user might ask, while the answers must be representative of the type of response the user would expect to receive from their query while staying faithful to the source data. Furthermore, a fully automated QA generation system lacks human oversight of the generated data and therefore must incorporate automatic quality controls. These controls must test both objective metrics, as well as custom metrics that reflect the intended RAG use case within its domain. As a result, while many QA generation systems have been proposed for various use cases, no single system published to date can generate useful data for all domains.

\subsection{Question-Answer Generation}

Automatic generation methods for QA data have evolved significantly since the introduction of transformers, and the methods proposed continue to evolve as LLM capabilities improve. Before transformers, proposals have been made to enhance sequence-to-sequence models for question generation using methods such as sentence- or paragraph-level information encoding \cite{du2017learning}, modelling question generation using answer embeddings \cite{sun2018answer}, and matching the answer with the passage before question generation \cite{song2018leveraging}. Ultimately, these have shown little success at improving downstream question-answering or reading comprehension tasks \cite{shakeri2020end}. Transformer-based methods have relied on training or fine-tuning of question generators within different architectures, where the more promising approaches \cite{alberti2019synthetic, puri2020training, liu2020asking, bartolo2021improving} adopted a three-step process: (1) answer detection using a pre-trained QA model (alongside custom heuristics), (2) question generation using a fine-tuned LLM to create questions conditioned on a passage and answer, and (3) question filtering using a model fine-tuned for reading comprehension to score questions. \\

The three-step system showed more success on downstream tasks than previous sequence-to-sequence approaches \cite{shakeri2020end}. However, an issue with the approach is that by first selecting candidate answers, it limits the scope of the content that any one single question can be based on, resulting in factual-style questions. It also provides less customisation optionality for the types of questions that are most important to target from a particular domain. Therefore, while these factual-style questions may be useful for benchmarking different RAG architectures or training basic RAG systems, they are less useful for testing a RAG system for a particular domain use case. Furthermore, the three-step system only generates free-text answers, which rely on semantic similarity metrics such as BLEU and ROUGE to be evaluated. A more concrete way of testing the RAG's knowledge is by asking it multiple-choice questions, where there is only one correct answer, and measuring its accuracy. Finally, multi-stage model training and fine-tuning is computationally expensive due to large transformer networks, and leaves room for errors to compound from stage to stage. \\

More recent methods have leveraged large generative models for synthetic data generation \cite{guo2024generativeaisyntheticdata}, including multi-lingual QA generation \cite{han2023multilingual}. Frameworks involving distillation and self-improvement, such as this one by DeepEval \cite{deepeval_qa_gen} and MLflow \cite{mlflow}, have been developed to generate questions and answers based on contexts using LLMs. These systems use simple prompts to ask the LLM to create a question based on a context paragraph \cite{mlflow}. The prompt includes instructions to ask a specific question and generate the answer using as much information as possible. However, these prompts are simple and generic, providing no instruction on what aspects of the context to focus on in the question, what specific details the question should include, or what the structure of the question should be. These issues make them unsuitable for generating domain-specific questions. Nevertheless, there is little to no literature on how to adapt LLM question generation methods to the requirements of a particular domain or generate diverse questions for testing various RAG abilities required for domain-specific question-answering. As such, this project utilises the MLflow \cite{mlflow} framework, adapted for the sustainability reporting domain, as a baseline to develop more sophisticated LLM question-generation methods.

\subsection{Question-Answer Evaluation}

Synthetic QA data generation techniques often employ an `overgenerate and filter' approach to improve question quality \cite{puri2020training}. Many evaluation metrics and methodologies exist to filter questions, and the choice and combination of evaluation metrics often depend on the specific use case of the RAG system and the characteristics of the target domain. One evaluation method is round-trip consistency, proposed by \cite{alberti2019synthetic}. This method involves using a question-answering model to predict an answer to a generated question, and then comparing this predicted answer to the original answer used to generate the question. This approach helps ensure that the generated questions are answerable and closely related to the intended answer. \\

Other fundamental aspects of QA evaluation include the relevance of questions (and answers) to the context as well as their factual accuracy. Diversity of questions is another key factor as different question structures and topics test different abilities of the RAG system. Nevertheless, little work has been done in using LLMs to evaluate other LLM-generated questions.

\section{Related Work in LLMs for Domain-Specific \\Question-Answering}

Early methods for developing domain-specific question-answering systems involve training RAG architectures, or various components of the architecture, from scratch. These techniques embed knowledge directly into the language models underpinning the system. REALM \cite{guu2020realm} was an early system that pre-trained specific knowledge (in this case, Wikipedia) into the retriever. This was done by augmenting the pre-training with a latent knowledge retriever, allowing the model to access documents during pre-training, fine-tuning, and inference. A pre-training approach is also adopted in RETRO \cite{borgeaud2022improvinglanguagemodelsretrieving}, which uses training and fine-tuning of autoregressive models conditioned on retrieved document chunks. Two recent innovations are RAG-end2end \cite{siriwardhana2022improvingdomainadaptationretrieval}, which explores training the retriever and generator jointly to improve their adaptation to domain-specific question-answering, and RAFT \cite{zhang2024raftadaptinglanguagemodel}, which trains the model to ignore `distractor' documents that don't help in answering the query. \\

The adoption of the above methods to real-world application is hindered by their computationally intensive and inflexible nature. It takes a large amount of external knowledge and significant model adaptation to train or fine-tune models from scratch. Furthermore, once embedded, it is difficult to keep the information up-to-date over time, as that would require re-training \cite{gao2024retrievalaugmented}. As such, modern domain-specific applications of RAG use pipelines where knowledge is retrieved from the external database rather than being embedded into the LLM parameters. This method is less computationally intensive and easier to maintain over time, prompting widespread adoption of LLMs and RAG systems to a wide range of domains. A particularly popular domain is finance, where human/expert-annotated datasets like FinQA \cite{chen2022finqa} and Financebench \cite{islam2023financebench} have enabled the development of large-scale systems such as BloombergGPT \cite{wu2023bloomberggptlargelanguagemodel}. Other popular domains include medicine \cite{xiong2024benchmarkingretrievalaugmentedgenerationmedicine} and pharmaceuticals, where the latter even explored RAG for question-answering in pharmaceutical regulatory compliance \cite{kim2024ragqaragintegratinggenerative}. \\

Nonetheless, there is no standard approach utilised across all domains, as the requirements of specialised systems differ across different domains and tasks. Furthermore, the development of high-quality question-answering systems is more difficult in the absence of a dataset that can be used for training, fine-tuning, and evaluating the system. As a result, this project presents a novel implementation of LLMs and RAG to develop and evaluate a chatbot for the sustainability reporting domain.

\section{Legal, Social Ethical and Professional Considerations }

The domain-specific RAG system presented in this project is crafted to be truthful and trustworthy. This is done by taking steps to minimise hallucinations through sourcing all information, providing succinct answers, and asking guiding questions to the user where more details are necessary to answer a query accurately. Nevertheless, hallucinations may only be minimised within the limits posed by the current technological state of large language models. Therefore, while the system is a reliable source of guidance most of the time, due to the nature of the LLMs underpinning the assistant, there still exists the risk of misrepresentation of the source content through hallucination. In the context of assisting companies in preparing regulatory reports, such hallucinations could potentially result in regulatory violations and legal consequences for the companies that rely on the hallucinated content. It is therefore advisable to leverage the system as purely a guiding assistant rather than as the single source of truth on sustainability reporting standards, and to verify information against the official IFRS reports. \\

The data used to augment the LLM - the IFRS Sustainability Reporting Standards - is not under copyright and may be used by companies freely to aid in the preparation of their regulatory reports. However, the data that the LLMs are trained on may contain information from unverified online sources, potentially without regard towards factuality, copyright, or social and political issues like discrimination, prejudice, and other biases. There may also exist the risk of misuse of LLMs in harmful or dangerous ways. To address these issues, this project takes steps to constrain the system to its specific domain by constructing an agent which ensures that unrelated queries are not answered, basing all responses on the source context, and constraining the language and vocabulary to the style of the source documents. Furthermore, providers of the LLMs used in this project - OpenAI, Anthropic, and Meta - have taken many measures to ensure their models are safe to use.\\

Finally, there exist the social and ethical concerns regarding the environmental impacts of the use of LLMs, which consume large amounts of energy during training on billions of parameters. The providers of these LLMs do not provide a guarantee that they use clean energy sources to train their models, and so reliance on them may perpetuate the use of non-renewable and/or polluting energy sources. However, there is currently no alternative for training LLMs in a less energy-intensive way, and efforts must be made on a large scale within AI to mitigate environmental impacts by using more clean and renewable energy sources.

\chapter{Synthetic Question-Answer Dataset }

This chapter presents the data and methods used to arrive at the final question-answer (QA) generation pipeline that is used to create a synthetic dataset of 1,063 QA pairs for evaluation of the RAG-based knowledge assistant. We start by introducing the domain and presenting the data collection and preprocessing pipeline. Next, we describe the methodologies adopted to generate and filter questions. First, we outline the requirements of our synthetic QA dataset and present the dataset structure created to meet these requirements. Then, we motivate our QA generation approach based on existing techniques, and describe the iterative development of the LLM architectures used, leveraging prompt engineering techniques like Chain of Thought (CoT) reasoning and Few-shot learning. Subsequently, we present a custom evaluation framework for QA quality, using both standard and novel metrics, and introduce three post-processing functions to improve the quality of the dataset. Finally, we present the results of our different question generation methods on the evaluation framework, define the final QA generation pipeline, and present a new dataset of QA pairs to be used for RAG training and evaluation in the sustainability reporting domain.

\section{Data collection and preparation}\label{data_collection}

This work adopts Sustainability Reporting as the knowledge domain, using data published by the International Financial Reporting Standards (IFRS) \cite{ifrsorgc9:online}, which provides international sustainability-related standards used by over 29,000 listed companies globally \cite{ifrs_companies}. These standards undergo periodic review and have recently been modified from previous versions in an effort to unify global reporting standards. The standards comprise 72 documents over 975 pages as described below:

\begin{enumerate}
    \item \textit{IFRS S1 General Requirements for Disclosure of Sustainability-related Financial Information (with an additional `accompanying guidance' document)}. This is the baseline financial-related reporting standard with the objective to ``require an entity to disclose information about its sustainability-related risks and opportunities that is useful to primary users of general purpose financial reports in making decisions relating to providing resources to the entity" \cite{ifrs_s1}. 
    \item \textit{IFRS S2 Climate-related Disclosures (with an additional `accompanying guidance' document)}. This is the baseline climate-related reporting standard with the objective to ``require an entity to disclose information about its climate-related risks and opportunities that is useful to primary users of general purpose financial reports in making decisions relating to providing resources to the entity" \cite{ifrs_s2}.
    \item \textit{Appendix B—Industry-based Disclosure Requirements}. This is the appendix to IFRS S2, comprising 68 separate industry-specific documents grouped into 11 different industry categories. These documents ``set out the [draft] requirements for identifying, measuring and disclosing information related to significant climate-related risks and opportunities associated with particular industries" \cite{ifrs_appendix}. It is important to note that the industry classification of companies is not always clear - IFRS states that entities can be ``associated with an industry through [their] business model, economic activities, production processes or other actions" \cite{ifrs_appendix}.
\end{enumerate}

The documents were downloaded in PDF format from the IFRS website at \cite{ifrs_s1,ifrs_s2,ifrs_appendix}, and the reports were manually examined to understand their structures and content. It was noted that, while S1 and S2 have custom structures, all Appendix reports follow the same structure. They start with identical title, disclaimer, and introduction pages, which are deleted from all but one report to avoid repetition. The material content in each report starts with an `Industry Description' section, followed by a `Sustainability Disclosure Topics \& Metrics' section. The latter contains one or two tables: all reports contain `Table 1. Sustainability Disclosure Topics \& Metrics', and most, though not all, contain `Table 2. Activity Metrics'. The columns of these tables are standardised across all reports, displaying the topics, with respective metrics, that must be reported for the particular industry, as well as as metric codes, categories, and units of measurement. The rest of the report contains detailed information on the topics and metrics outlined in the tables, structured as one section per topic, with subsections pertaining to each topic metric. \\

Finally, it is important to note that, having recently undergone modifications from previous versions, the standards contain many sections of crossed-out text that have been removed from prior versions, often followed by sections of underlined text that have been added. Crossed-out and newly added text is present in both tables as well as free-text sections. As outlined in Section \ref{data_processing}, crossed-out text is removes from the data, and underlined text is kept.

\subsubsection{Data preprocessing} \label{data_processing}

\begin{figure}[H]
    \centering
    \includegraphics[width = 0.8\textwidth]{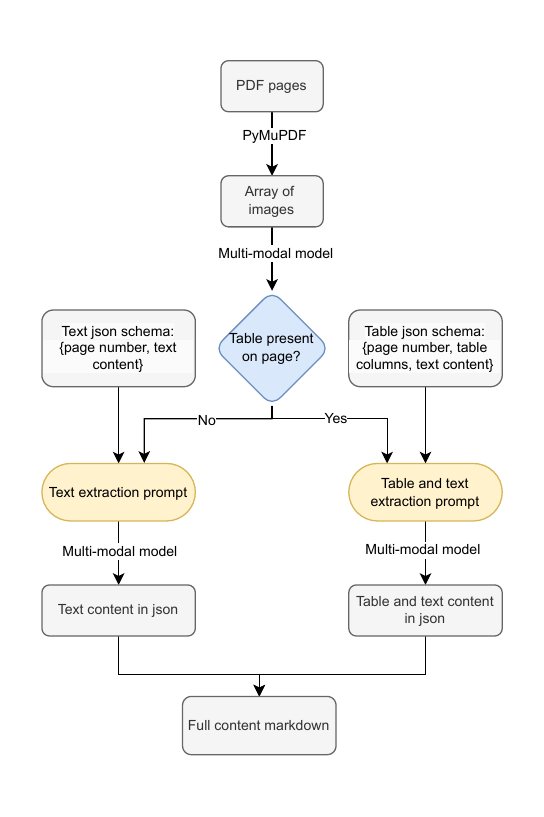}
    \caption{Multi-modal PDF to markdown parsing pipeline.}
    \label{fig:pdf_parse}
\end{figure}

Leveraging the structured nature of the IFRS reports, a custom multi-modal pipeline (Figure \ref{fig:pdf_parse}) was designed to parse the text and table content into markdown format, preserving the hierarchy of report sections. A multi-modal model was chosen for parsing PDFs as it can accurately differentiate between crossed-out and underlined text chunks in the reports, and only extract the underlined ones. Other PDF parsing packages and OCR techniques were tested and were unsuccessful at accurately extracting only the underlined text and not the crossed-out text. The prompt used for text and table extraction is displayed in Appendix \ref{lst:pdf_parse_prompt}. The prompt is designed with particular attention towards crossed-out text, as well as to preserve the hierarchical structure of the report sections. \\

The multi-modal parsing pipeline follows four steps:

\begin{enumerate}
    \item PDF to image conversion: Each page of the PDF is converted into an image using a zoom factor of 2.5 to increase the resolution of the images for more accurate content parsing. The PyMuPDF package is used for this step.
    \item Table presence check: Each image is passed through a multi-modal model which checks if a table is present on the page. The model used is Claude 3.5 Sonnet.
    \item Conditional processing: Pages without tables are inserted into a text extraction prompt which, combined with a JSON schema for structuring the output, is passed through a multi-modal model to extract the text content and associated page number into a JSON file. Pages with tables present are inserted into a table and text extraction prompt, along with a more complex JSON schema, to extract tabular data with custom table columns (alongside any other text content and the page number). 
    \item JSON to markdown conversion: The JSONs for each page are combined into a structured markdown file that maintains section headers and hierarchies, and isolates the tables from the free-text content.
\end{enumerate}

The resulting 72 markdowns are saved to be used in QA generation as well as in the RAG system development described in Chapter 4.

\newpage
\section{Methods}

\subsection{QA design}\label{qa_design}

The ultimate aim of the knowledge assistant is to answer user queries correctly based on the source data. To ensure this, the RAG system must be evaluated on questions that are (1) representative of real user queries and (2) useful for testing the capabilities of the RAG system. The QA dataset is designed based on these two principles.\\

To design questions that are representative of real-life queries, several sources were used. First, the contents of the IFRS reports were manually analysed to identify key topics that are often repeated or heavily emphasised, as well as smaller but critical details that a person likely wants to know. Second, a range of real corporate sustainability reports from SASB (now IFRS) reporting companies were downloaded from \cite{sasb_reports} and studied to identify the ways in which companies use IFRS information to prepare sustainability reports. In most cases, companies prepare tables that explicitly provide the metrics required by IFRS, meaning they are very likely to ask detailed questions about metrics. Third, inspiration for typical user queries was taken from FinQA \cite{chen2022finqa} and Financebench \cite{islam2023financebench} - two prominent expert-annotated QA datasets that are based on corporate reporting. Both the questions themselves as well as their preparation methods were analysed. From the actual Financebench and FinQA question datasets, 20 questions were selected and adapted to the Sustainability Reporting domain. From the Financebench preparation methodology, the idea of preparing ``domain-relevant questions" that are ``generically relevant" to the user's task was adopted. In the case of Financebench, the user's task is defined as the financial analysis of a publicly traded company \cite{islam2023financebench}, whereas the user's task defined for this project is the use of IFRS standards to prepare corporate sustainability reports. It is important to note that these methods were used by experts to manually create questions, whereas this project adapts these methods to LLMs using prompt engineering. \\

Based on these sources, a list of 10-18 user-representative sample questions was created for each question category - Multiple-choice (MCQ) Local, MCQ Cross-industry, Free-text Local, Free-text Cross-industry - to be used for few-shot prompting techniques in the question generation pipeline as discussed in Section \ref{QA_gen}. Six examples are displayed in Listing \ref{lst:qa_structures_example} below, and the full list of sample questions can be found in Appendix \ref{lst:qa_structures}.

\begin{lstlisting}[language=JSON,firstnumber=1,label={lst:qa_structures_example},caption={Select user-representative question structures}]
Local single-hop:
"Can you provide for me the unit of measure I should use for the xxx metric in the xxx industry?",
"On what page can I find details about xxx for the xxx industry?"

Local multi-hop:
"Can you provide for me the unit of measure I should use for each of the metrics for 'xxx' topic(s) in the xxx industry?",
"What is the category of xxx metric in the xxx industry and how should this metric be calculated?"

Cross-industry single-hop:
"What topics should I report on for xxx and xxx (and xxx...) industries?",
"Give me all the metrics from Table 1 that are in the category 'Discussion and Analysis' for xxx and xxx (and xxx...) industries."

Cross-industry multi-hop:
"Give me the codes for all the 'Quantitative' metrics in Table 1 for xxx and xxx (and xxx...) industries.",
"How does the calculation of xxx-related metrics differ between xxx and xxx (and xxx...) companies?"
\end{lstlisting}

To design questions that are useful for testing the RAG system, questions were generated on three dimensions: question span, question complexity, and answer structure.\\

\textbf{1. Question span:} testing the RAG's ability to retrieve information from and reason over multiple source documents. The options are:
\begin{itemize}
    \item Single industry (`Local'): Questions that cover only one industry-related source document. They are designed to test retrieval abilities from one document.
    \item Two industries (`Cross-industry'): Questions that cover a pair of industries, in a way that compares/contrasts or gathers information from both industries. Industry pairs were chosen by an LLM prompted to pair up industries that would realistically be compared based on the industry descriptions. These form a key part of the evaluation dataset, as many IFRS reporting companies span different industries, and are designed to test the RAG's ability to identify the correct documents and source the relevant chunks from both.
\end{itemize}

\textbf{2. Question complexity:} testing the RAG's ability to reason through a question and retrieve necessary data. The options are:
\begin{itemize}
    \item Single-hop: Questions where the answer can be extracted directly from the document. They are designed to evaluate the RAG system's ability to answer factual questions based on small snippets of the source document.
    \item Multi-hop: Questions that require information from multiple sections of one or more reports, or involve multiple logical steps, to be answered. They are designed to evaluate the RAG system's ability to reason in multiple steps and extract multiple information chunks to answer a question.
\end{itemize}

\textbf{3. Answer structure:} testing the RAG's knowledge as well as its ability to answer a question correctly and coherently. The options are:
\begin{itemize}
    \item Multiple-choice: Questions with five answer options and only one correct answer option. They are designed to test the accuracy of the RAG system in an objective way that does not rely on NLP metrics.
    \item Free-text: Questions that have a free-text answer. They are designed to test the accuracy and coherence of the RAG system using traditional NLP metrics such as BLEU and ROUGE.
\end{itemize}

These categories are combined into eight different question configurations: local single-hop multiple-choice questions (MCQ), local multi-hop MCQ, local single-hop free-text questions (FTQ), local multi-hop FTQ, cross-industry single-hop MCQ, cross-industry multi-hop MCQ, cross-industry single-hop FTQ, and cross-industry multi-hop FTQ.

\subsection{QA Generation}\label{QA_gen}

The overall aim of this project is to generate questions that are as specific as possible to minimise the chance of hallucination, and then generalise the phrasing of the questions in post-processing. As discussed later in this chapter, LLMs tend to generate vague questions based on a given context. As such, steps are taken to guide the LLM to create questions and answers that are highly specific to the domain to maximise their faithfulness and relevance to the text. Post-processing functions are then designed to enable the adjustment of highly specific questions to better mimic natural human questions while maintaining their underlying materiality.\\

The question generation and evaluation pipeline is defined in Algorithm \ref{alg:final-question-generation}. The steps of the pipeline are as follows:
\begin{enumerate}
    \item For each question setting (or type) described in Section \ref{qa_design}, a certain number of questions are generated using LLM prompting. Three techniques are tested for this step: a baseline prompt, a Chain-of-Thought (CoT) prompt, and a prompt that uses both CoT and provides few-shot examples of sample questions.
    \item Each question is then evaluated using a number of custom metric implementations. The evaluation metrics designed and employed are faithfulness, relevance, and specificity. Free-text questions are also evaluated using BLEU and ROUGE-L.
    \item Questions that fail to pass certain evaluation thresholds may be improved in post-processing using a quality improvement function.
    \item Questions that are ``too" specific may be passed through a generalisation function to make them sound more vague while maintaining their factfulness.
    \item A similarity filter is used to exclude question duplicates.
    \item MCQs are checked to ensure they have a `single best answer' (SBA). Questions that fail this test are discarded.
\end{enumerate}
\begin{algorithm}[H]
\caption{Final Question Generation Pipeline}
\label{alg:final-question-generation}
\begin{algorithmic}[1]
\Require Set of industry markdowns $M$, Number of questions $N$, Question types $T$ (free-text, MCQ)
\Ensure Set of quality question-answer pairs $Q$
\State $Q \gets \emptyset$
\For{each question type $t \in T$}
    \For{$i = 1$ to $N$}
        \State Select $k \geq 1$ industry contexts $C = \{c_1, \ldots, c_k\}$ from $M$
        \State $q \gets \Call{GenerateQAPair}{C, t}$ 
        \Comment{using Claude 3.5 Sonnet}
        \State $metrics \gets \Call{EvaluateMetrics}{q}$
        \If{$\exists m \in metrics : m < \theta_m$} \Comment{$\theta_m$: metric threshold}
            \State $q \gets \Call{QualityImprovement}{q, metrics}$
        \EndIf
        \If{$q$ contains industry name}
            \State $q \gets \Call{QuestionGeneralization}{q}$
        \EndIf
        \State $is\_similar \gets \Call{SimilarityFilter}{q}$
        \If{$\lnot is\_similar$ and $\forall m \in metrics : m \geq \theta_m$}
            \If{$t$ is multiple-choice}
                \State $has\_single\_best\_answer \gets \Call{SingleBestAnswer}{q}$
                \If{$has\_single\_best\_answer$}
                    \State $Q \gets Q \cup \{q\}$
                \EndIf
            \Else
                \State $Q \gets Q \cup \{q\}$
            \EndIf
        \EndIf
    \EndFor
\EndFor
\State \Return $Q$
\end{algorithmic}
\end{algorithm}
To design the prompt architecture for QA generation, three approaches were tried. Firstly, a baseline prompt was used to generate questions based on the markdown content as a baseline. To reduce the chance of hallucination, a second approach was adopted using a more sophisticated prompt with Chain-of-Thought (CoT) reasoning. Finally, to ensure that the questions are representative of real user queries, the last prompt-based approach incorporates few-shot learning into the prompt. \\

All questions are structured to have the question, the answer, the reference text that the question and answer is based on, and the pages where the reference text is sourced from. Multiple-choice questions also have answer options A through E, where only one option is the correct answer. 

\subsubsection{Baseline Prompting}

In the baseline method, question-answer pairs are generated with an LLM function call using a prompt that gives the LLM context of the task at hand - generating questions based on the markdown from the perspective of a company preparing sustainability reports. However, it is given no guidance on how these questions should be generated. The prompt takes as inputs the markdown and the question complexity: single-hop or multi-hop. For questions comparing different industries, the prompt asks the LLM to generate questions based on all markdowns given. The prompt used is shown in Listing \ref{lst:naive_prompt}.\\
\begin{lstlisting}[language=JSON,firstnumber=1,label={lst:naive_prompt},caption={Baseline prompt for `local' multiple-choice question generation}]
system = "You are a sustainability reporting expert that helps companies draft their corporate sustainability reports using the IFRS reporting standards. You are preparing some questions that a company might ask while preparing its sustainability report, for which the answer can be taken from the context given in the markdown below."

user =  f"""
Here is the markdown content:
{markdown_content}

Based on the markdown content, generate {n} multiple-choice questions of type {qa_type}.

Generate {n} QA pairs for the specified type and return them using the provided schema."""
\end{lstlisting}

LLM `tool use' is used to ensure the output is in the required format containing the question, answer, reference text, and pages (and answer options for MCQs). A JSON schema is used as the tool to structure the output, and this is shown in Appendix \ref{lst:mcq_schema}.

\subsubsection{Chain-of-Thought Prompting}

In the second approach, several additions are made to the prompt to reduce the possibility of hallucination and improve the style of questions generated. The main addition is CoT reasoning to guide the LLM on the steps to take to generate questions, adapting traditional question-generation methods. Specifically, the LLM is prompted to:

\begin{enumerate}
    \item Select a list of one or more sentences/snippets of the markdown content that can be used to form an answer to a question. This will form the reference text.
    \item Write a question that requires the reader to understand the content of the selected text to answer correctly. The question should be based only on the selected text and should not require any additional information.
    \item Write five answer options, one of which is correct and the other four are incorrect. The correct answer should be complete and taken verbatim from the selected section(s) of the markdown content.
\end{enumerate}

This process adapts the traditional question-generation methods to LLMs by first guiding it to select the text that serves as the response before generating the QA pair. This adaptation has the advantage that the LLM has full context of what the text is about and is thus able to jointly select pieces of reference text that are not only most relevant to the specific domain but also would make the ``best" questions. Furthermore, it can do so with a reinforced understanding of its role as a human that wants information specifically for a company preparing sustainability reports. \\

Further additions to the prompt for MCQs include making explicit that the questions should have a single best answer, which should require very specific information from the context to be answered, as well as ensuring that the other answer options are not so obviously wrong that they can easily be discarded. The prompt used to achieve this (for local multiple-choice questions) is shown in Listing \ref{lst:cot_prompt}. \\

\begin{lstlisting}[language=JSON,firstnumber=1,label={lst:cot_prompt},caption={CoT prompt for `local' multiple-choice question generation}]
system = [unchanged]

user = f"""
Here is the markdown content:
{markdown_content}

Based on the markdown content, generate {n} 'Single Best Answer' (SBA) questions that have only one correct answer out of five options. The correct answer should not be obvious and *should really require specific information from the source document to be able to be answered*. The incorrect answer options should not be so ridiculous or extreme that they are obviously wrong. The questions must be of the type: {qa_type_str}

The questions should be ones that could occur when a human wants information from the chatbot. They should be directly relevant to companies preparing their sustainability reports and reflect real-world scenarios that reporting teams might encounter.

To generate questions, follow these steps:
1. Select a list of one or more sentences/snippets of the markdown content that can be used to form an answer to a question. This will form the reference text. Remember this should be relevant to the human for drafting sustainability reports.
2. Write a question that requires the reader to understand the content of the selected text to answer correctly. The question should be based only on the selected text and should not require any additional information. Remember this should be the type of question a human would ask when drafting sustainability reports.
3. Write five answer options, one of which is correct and the other four are incorrect. The correct answer should be complete and taken verbatim from the selected section(s) of the markdown content.

Generate {n} unique and diverse QA pairs for the specified type and return them using the provided schema.
"""
\end{lstlisting}

\subsubsection{CoT + Few-shot Prompting}

The third approach introduces few-shot examples to the prompt using the sample questions curated from different sources as described in Section \ref{qa_design} (the list of sample questions can be found in Appendix \ref{lst:qa_structures}). The LLM is given 10-12 question styles and asked to follow the CoT question generation steps to create questions in the same style. It is asked to select the question styles at random and adapt any that do not directly make sense in the context of the provided markdown(s).

\begin{lstlisting}[language=JSON,firstnumber=1,label={lst:memprompt},caption={CoT + few-shot prompt for `local' multiple-choice question generation}]
system = [unchanged]

user = f"""
[unchanged]

Some example {qa_type} question structures are shown below. Please choose {n} structures at random but do not limit yourself to these types only. If some of these structures do not make sense given the content of the document, adapt them to the context as appropriate or choose ones that you think are appropriate.

Here are some question structures:
{question_structures_str}  
    
Generate {n} unique and diverse QA pairs for the specified type and return them using the provided schema.
"""
\end{lstlisting}

All of the prompting techniques were experimented with using four different temperature settings - 0.0, 0.2, 0.5, 1.0 - to test the impact of varying LLM creativity on the questions generated.

\subsection{QA Evaluation}

\subsubsection{Faithfulness and Relevance}

After the QA pairs are generated, they are evaluated on three LLM-based metrics - faithfulness and relevance, as well as a custom `domain specificity' metric. Faithfulness is a measure of how accurately LLM-generated text is grounded in the provided context(s) \cite{es2023ragas}. A faithful question must be directly answerable from the given information without any need for external knowledge or inference, while a faithful answer is true based on the context. Relevance is a measure of how focused LLM-generated text is on the given content, containing as little irrelevant information as possible \cite{es2023ragas}. A relevant question precisely aligns with the specific context taken from the report, while a relevant answer addresses the actual question that was given. \\

Implementing QA evaluation is a more complex task than a simple evaluation of LLM-generated text since questions and answers that are not simply factual typically deviate in phrasing and/or semantics from the source document - this can be seen in the drastic drop of BLEU and ROUGE-L scores as questions span more industries (see Table \ref{tab:perindustry_results}). Typical LLM-based checks of faithfulness and relevance are designed to evaluate LLM-generated responses to queries, rather than the quality of a whole question-answer pair relative to a complete source document, making them unsuitable for a direct application to the QA evaluation task. This was tested using two popular open-source evaluation frameworks - LlamaIndex \cite{llamaindex_eval} and DeepEval \cite{deepeval} which use various prompting techniques, most notably chain of thought, to measure metrics on a binary (LlamaIndex) or continuous (DeepEval) scale. They were manually tested on a subset of questions and did not prove to be effective at identifying bad question quality (see Appendix \ref{lst:opensource_eval} for an example). The tools perform particularly poorly on evaluating the relevance of a question, as they rely on generic prompts that do not provide context on how to measure relevance as defined from the perspective of a company using the context to produce sustainability reports. Nevertheless, these tools are effective at evaluating simple (non-question) statements relative to source context \cite{deepeval_paper}. \\

\begin{figure}[H]
    \centering
    \includegraphics[width = 0.9\textwidth]{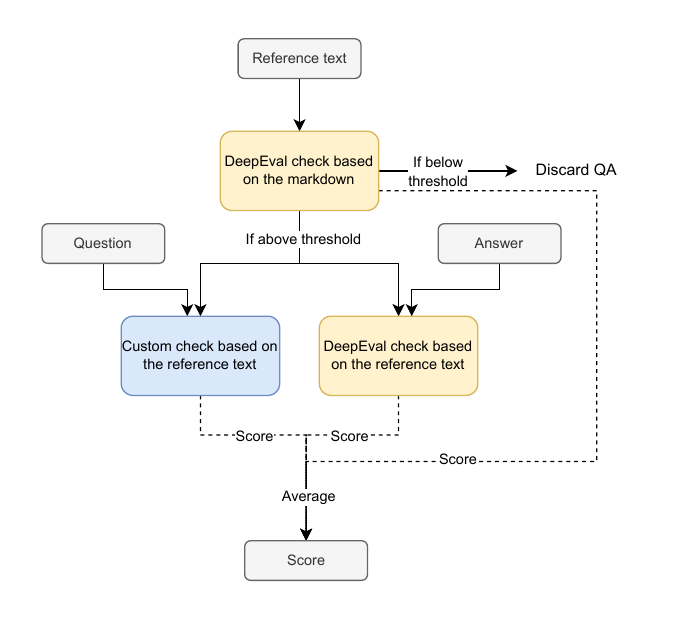}
    \caption{Faithfulness and relevance QA evaluation pipeline}
    \label{fig:metrics_arch}
\end{figure}

This project combines open-source evaluation tools with a custom `question quality' evaluator into an evaluation pipeline shown in Figure \ref{fig:metrics_arch}. With this, the resulting tools are (1) able to evaluate full question-answer pairs and (2) targeted to the sustainability reporting domain. DeepEval (chosen as it provides scores on a continuous scale) is used to evaluate the faithfulness and relevance of the reference text relative to the entire report markdown, as well as of the answer relative to the reference text. A standard evaluation tool is suitable for these tasks because it evaluates text taken straight from the document without the need for additional context.\\

A custom evaluator is used to check the quality of the question itself relative to the reference text. This evaluator uses domain context and few-shot prompting to guide the LLM on grading the faithfulness and relevancy of the question. The custom evaluator used is shown in Listing \ref{lst:fewshot_prompt}. Note that, for relevance, the LLM is also asked to check whether the question relates to all the industries it is set to cover (e.g. if it is asked to create a question comparing two industries, but the question asks about only one of those industries, it is considered irrelevant). \\

A design choice is made to automatically exclude any QA pairs that have a reference text faithfulness or relevance below 0.7, to avoid falsely inflating the scores of the question and answer that are based on the reference text. This threshold was chosen from manual inspection of question quality to define what is generally deemed acceptable. The average faithfulness/relevancy score for the reference text, question, and answer is then taken. Note that the custom question evaluator works on a scale from 1 to 10 as, experimentally, this has helped the LLM be more nuanced in its scoring than using a 0-1 scale. Therefore, the DeepEval scores are re-based on a 1-10 scale before taking the average. \\

\begin{lstlisting}[language=JSON,firstnumber=1,label={lst:fewshot_prompt},caption={Prompt for question faithfulness/relevance evaluation}]
system = "You are an expert question evaluator. Given a question, you assess its faithfulness and relevance relative to the reference text.

user = f"""
Critically evaluate the following question based on the provided context for the relevant {'industry' if len(industries) == 1 else 'industries'}. Be extremely rigorous and unforgiving in your assessment.

Faithfulness: Measure how accurately the question is grounded in the provided context{'s' if len(industries) > 1 else ''}. A faithful question must be directly answerable from the given information without any need for external knowledge or inference. Be extremely critical - even minor discrepancies or omissions should significantly impact the score.

Relevancy: Assess how precisely the question aligns with {'the' if len(industries) == 1 else 'each'} industry's specific context, challenges, and goals. A highly relevant question should directly address key aspects, metrics, or challenges unique to the industry. Be very strict - even slight deviations from industry-specific concerns should result in lower scores. ALSO: if the question does not cover all the industries required, it must be considered irrelevant and given a score of 1.

Score both metrics on a scale of 1 to 10, where:
1-2: Completely irrelevant or unfaithful
3-4: Major flaws in relevancy or faithfulness
5-6: Moderate issues, but still lacking
7-8: Generally good, with minor issues
9-10: Excellent, near-perfect alignment

Example Context:
"The Apparel, Accessories & Footwear industry faces significant sustainability challenges, particularly in raw materials sourcing. Key metrics include:
1. Percentage of raw materials third-party certified to environmental and/or social sustainability standards.
2. Priority raw materials: Description of environmental and social risks and/or hazards associated with priority raw materials used for products.
3. Environmental impacts in the supply chain: Percentage of (1) Tier 1 supplier facilities and (2) supplier facilities beyond Tier 1 that have completed the Sustainable Apparel Coalition's Higg Facility Environmental Module (Higg FEM) assessment or an equivalent environmental data assessment."

Examples:
Good Question  (Faithfulness: 10, Relevancy: 9):
"What percentage of raw materials in the Apparel, Accessories & Footwear industry should be third-party certified to environmental or social sustainability standards, according to the context?"
Explanation: This question directly addresses a specific metric mentioned in the industry context and can be answered solely based on the provided information.

Very Bad Question (Faithfulness: 1, Relevancy: 2):
"What is the average salary of a fashion designer in New York City?"
Explanation: This question is neither relevant to the industry's sustainability metrics nor answerable from the given context.
"""

\end{lstlisting}

\subsubsection{Domain Specificity}

Domain specificity is a newly introduced metric in this project to measure how specific the QA pair is to the sustainability reporting domain through the use of key concepts or trends found in the report. The question "What does Table 1 contain?" is generic, whereas "Can you provide me the unit of measure for the \{xxx\} metric in Table 1?" is more specific, as it gives more context for what the question should focus on from a domain perspective. Maintaining high question specificity ensures that the generated questions are representative of the domain and reduces the possibility of hallucination by guiding the LLM to the exact details it should be using from the source document to generate the question and answer. Furthermore, specific questions are less likely to have obvious answers that can be determined without reference to the source document (i.e. without RAG), so they are more useful for testing the system's true domain knowledge. \\

Domain specificity is measured on the question (plus answer options if it is an MCQ) relative to the whole markdown report (rather than just the reference text). This gives the LLM full context on what the report is about rather than a single text snippet, enabling it to identify keywords or ideas that are important in the report and therefore should be mentioned in the question. A scale of 1-10 is used, and the few-shot prompt is shown in Listing \ref{lst:qa_eval_prompt}.

\begin{lstlisting}[language=JSON,firstnumber=1,label={lst:qa_eval_prompt},caption={Few-shot prompt for domain specificity}]
system: = ""You are an expert question evaluator."

user = f"""
    Full content:
    {full_context}

    Evaluate the specificity of the following question based on the provided context and compared to the highly specific question examples provided.
    Score the specificity on a scale of 1 to 10, where 1 is extremely broad or general and 10 is very specific. 
    
    {"Consider both the question itself and the answer options given." if is_multiple_choice else "Consider the question and the expected level of detail in the answer."}
    
    If a question requires a very specific answer directly from a specific sentence/part of the document, it is considered more specific. If a question can be answered in multiple ways or is broad, it is considered less specific.

    To help you, here are some example questions below:

    Highest specificity questions that score 10:
    "What is the unit of measure for the 'Percentage of raw materials third-party certified to an environmental and/or social sustainability standard, by standard' metric in the Apparel, Accessories & Footwear industry (as listed in the relevant table)?"

    High specificity questions that score 8:
    "What topics are covered in the 'Raw Materials Sourcing' section of the Apparel, Accessories & Footwear document, and what are the key takeaways for a company writing its sustainability report in this industry?"

    Medium specificity questions that score 6:
    "A company in the Household & Personal Products industry is facing water scarcity issues in multiple manufacturing locations. What is the most comprehensive approach to address this challenge in their sustainability report?"

    Low specificity questions that score 3:
    "How might the increasing focus on energy efficiency certifications in the appliance industry influence future regulatory trends and consumer behaviour?"

    Lowest specificity questions that score 1:
    "What broader implications does the industry's focus on energy management have for environmental sustainability?"

    Now the question to be evaluated:
    Question: {question_data['question']}
    """
\end{lstlisting}

\subsubsection{BLEU and ROUGE-L}

Finally, for free-text questions, BLEU and ROUGE-L are calculated for the answer relative to the reference text. ROUGE-L is chosen (over ROUGE-N) as it is designed to evaluate the semantic similarity and content coverage of text regardless of word order. ROUGE-N on the other hand simply evaluates the grammatical correctness and fluency of the generated text, which is not a requirement of the answers - succinct and factually accurate answers are preferred.\\

MCQ answers are not evaluated using BLEU and ROUGE-L. This is because low BLEU and/or ROUGE-L scores are not indicative of a wrong answer. For example, the correct answer option for an MCQ may be ``All of the above", which is not a phrase found in the text, and so will score 0 on both metrics despite being correct. An extension of this problem is that, if all answer options were to be evaluated instead of just the single correct one, multiple answer options may score highly as the options are designed to be similar enough that the correct answer is not obvious. It then becomes difficult to discern between correct and incorrect options from metric scores.

\subsection{Post-processing}

A selection of post-processing functions was developed to make the questions as high quality as possible (i.e. free of hallucination, relevant, and domain-specific) while making them sound human-like. For this project, human-like questions are defined as having some degree of vagueness, particularly with regards to the industry being asked about. To achieve this aim, two functions were introduced - quality improvement, and question generalisation. Additionally, MCQs are verified to ensure they have a single best answer (SBA) among the answer options. Finally, a similarity filter is introduced to remove identical question copies (in case these arise).

\subsubsection{Quality Improvement Function}

The first post-processing step involves passing questions that fail to meet a threshold on any of the three quality metrics through a quality improvement function that is designed to address the identified shortcomings. The flagged questions are passed along with their relevant industry markdown(s) through a prompt that instructs the LLM to slightly adjust the question on the required aspect. For example, if a question scored low on faithfulness, the prompt instructs the model to reformulate the question to more closely align with the factual information from the report. Similarly, for questions with low relevance scores, the prompt guides the model to focus more tightly on the specific sustainability reporting context from the report. This approach targets improvements to address each question's weaknesses while preserving its original intent and structure. The prompt used for this function is shown in Appendix \ref{lst:quality_improvement_prompt}.\\

To avoid any chances of hallucination by repeated prompting (as the LLM "forgets" the initial goal), this function is applied only once, and the question metrics are checked again. If the improved question still fails to meet the threshold, or meets the threshold for the original weak metric but now fails on another metric, it is discarded. Additionally, if the original question fails to meet the threshold on more than one metric, it is discarded, as it would not provide the LLM enough ``starting material" for the improvement function to reformulate the question while preserving its initial intent. \\

There is no generally accepted choice of threshold for faithfulness and relevance, as thresholds are designed to be adapted to the aim of the LLM application. In principle, any score above 0.5 would suggest that the LLM is better than random and may be deemed acceptable. For QA evaluation, the thresholds are determined based on the results of the experiments shown in Table \ref{tab:qa_evals_methods} of Section \ref{qa_experiment_results}.

\subsubsection{Question Generalisation Function}

The second post-processing function generalises questions that are ``too specific to be human". It alters any mentions of very specific industry names into more generic industry ``areas", mimicking the more natural conversational-style queries that a user would ask. For example, a question that originally asked ``What does the IFRS require companies to disclose regarding energy management for the Processed Foods industry?" might be rephrased to ``What does the IFRS require food companies to disclose regarding energy management?". This generalisation increases the difficulty of the questions, requiring the RAG system to have a deeper understanding of the content rather than relying on keyword matching. The efficacy of this function is tested qualitatively. The prompt used for this function is shown in Appendix \ref{lst:generalisation_prompt}.

\subsubsection{Single Best Answer Verification}

The questions are filtered to remove any MCQs that have more than one correct answer (i.e. that do not pass the SBA test). This is checked by giving a strong LLM (Claude 3.5 Sonnet) the full markdown context and asking it to choose all correct answer options based on it. If more than one option is chosen, the question fails the SBA test and is discarded. Finally, a similarity filter is created to remove identical questions. The prompt used for this function is shown in Appendix \ref{lst:sba_prompt}.

\subsubsection{Similarity Filter}

Finally, to avoid question repetition, the questions are passed through a similarity filter to remove identical questions. The nature of the question generation pipeline, which uses few-shot prompting to create very specific questions, means that many questions are likely to be very similar in their structures. As such, the filter is set to a threshold of 0.99\% similarity, and questions above this are discarded.

\section{Results}

\subsection{Question Embeddings}

\begin{figure}[H]
    \centering
    \includegraphics[width = \textwidth]{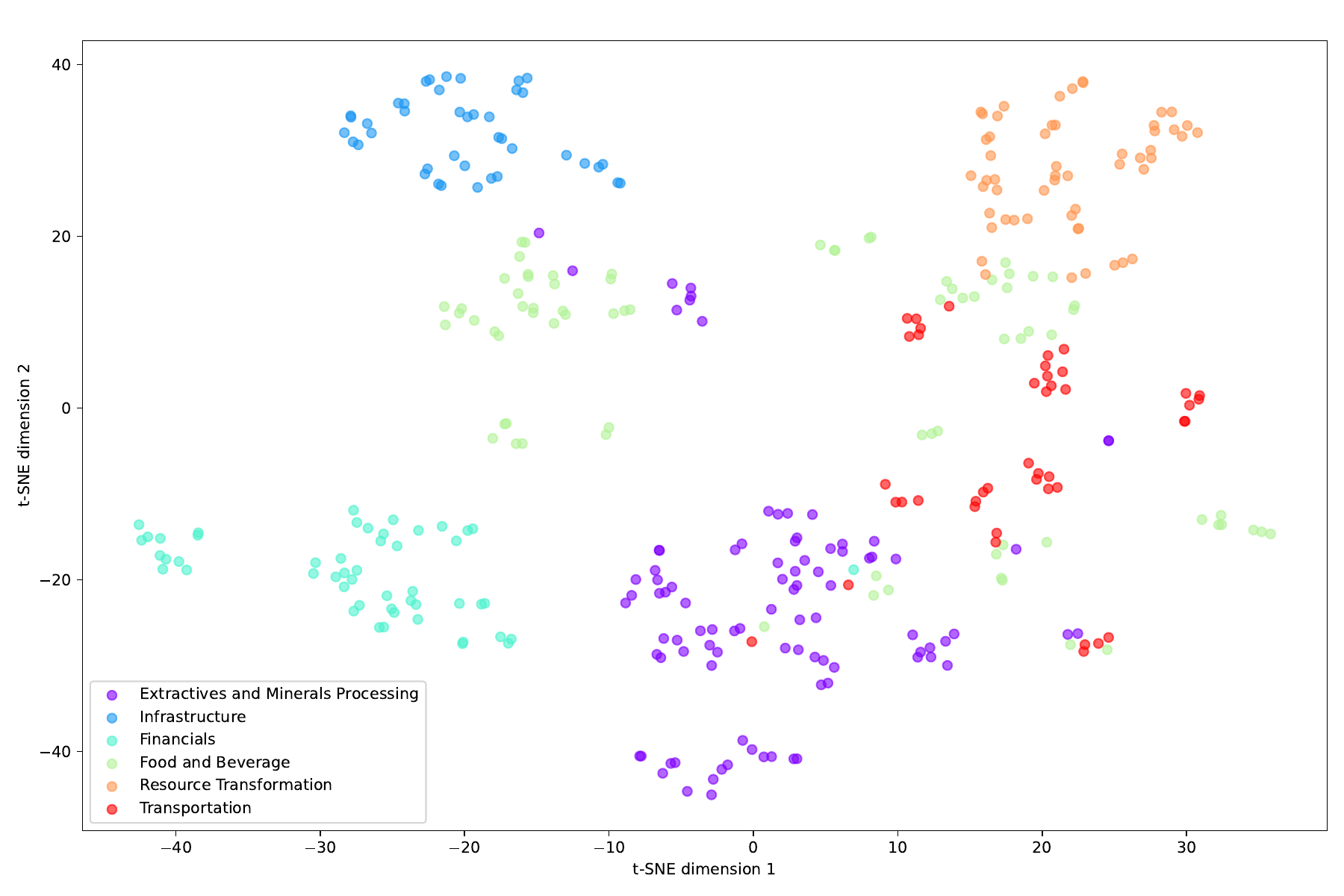}
    \caption{Sample t-SNE visualisation of question embeddings, coloured by industry group}
    \label{fig:question_embeddings}
\end{figure}

The embeddings for a sample of local free-text questions from six industry categories are mapped using t-SNE (t-distributed Stochastic Neighbour Embedding), a dimensionality reduction technique for visualisation of high-dimensional data. For each industry category, questions are randomly selected from each industry. The questions used are generated using the CoT + few-shot method. The results are shown in Figure \ref{fig:question_embeddings}.

\subsection{Experimental Results}\label{qa_experiment_results}

The three question generation methods outlined in Section \ref{QA_gen} are used to generate a sample of questions for quality evaluation experiments. For each method, the questions are generated according to the settings described in Section \ref{qa_design}:
\begin{itemize}
    \item Question span: For local questions, eight industries are chosen at random, while for cross-industry questions, eight industry pairs are chosen out of the LLM-generated industry pairs (generation of these pairs is described in Section \ref{qa_design}). (Note that a subset of industry (pairs) is used for experimentation due to cost reasons. The full QA dataset will be generated on all industries and industry pairs.)
    \item Question complexity: Single- and multi-hop questions are generated.
    \item Answer structure: MCQ and free-text questions are generated.
    \item Temperature: All settings are tried with four temperatures: 0.0, 0.2, 0.5, 1.0.
\end{itemize}

To conduct experiments (before running the final question generation pipeline), 32 questions of each type and temperature are created and evaluated. Tables \ref{tab:qa_evals_methods}, \ref{tab:qa_evals_methods_free}, and \ref{tab:temperature-comparison} below show the average scores of question evaluations across different methods and temperatures.

\begin{table}[H]
\centering
\begin{tabular}{lcccccc}
\hline
              & \multicolumn{3}{c}{MCQ Local}                                                & \multicolumn{3}{c}{MCQ Cross-industry}                                       \\ \cline{2-7} 
              & Baseline & CoT     & \begin{tabular}[c]{@{}c@{}}CoT +\\ Few-shot\end{tabular} & Baseline & CoT     & \begin{tabular}[c]{@{}c@{}}CoT +\\ Few-shot\end{tabular} \\ \cline{2-7} 
No. Questions & 64       & 64      & 64                                                       & 64       & 64      & 64                                                       \\
Faithfulness  & 6.17     & 7.25    & \textbf{9.45}                                                     & 5.58     & 6.06    & \textbf{7.69}                                                     \\
Relevance     & 6.25     & 7.78    & \textbf{9.68}                                                     & 3.46     & 5.93    & \textbf{7.86}                                                     \\
Specificity   & 5.61     & 7.47    & \textbf{9.24}                                                     & 5.04     & 6.31    & \textbf{7.69}                                                     \\
SBA (\%)      & 67.19\%  & 81.25\% & \textbf{92.19\%}                                                  & 68.76\%  & 75.01\% & \textbf{92.19\%}                                                  \\ \hline
\end{tabular}
\caption{Experimental results per method, averaged over single- and multi-hop MCQs. The experiment was done with questions generated on a selection of 8 industries (for local) and 8 industry pairs (for cross-industry) with a temperature of 0.5.}
\label{tab:qa_evals_methods}
\end{table}

\begin{table}[H]
\centering
\begin{tabular}{lcccccc}
\hline
              & \multicolumn{3}{c}{Free-text Local}                                       & \multicolumn{3}{c}{Free-text Cross-industry}                              \\ \cline{2-7} 
              & Baseline & CoT  & \begin{tabular}[c]{@{}c@{}}CoT +\\ Few-shot\end{tabular} & Baseline & CoT  & \begin{tabular}[c]{@{}c@{}}CoT +\\ Few-shot\end{tabular} \\ \cline{2-7} 
No. Questions & 64       & 64   & 64                                                       & 64       & 64   & 64                                                       \\
Faithfulness  & 5.68     & 7.12 & \textbf{8.75}                                                    & 5.09     & 5.89 & \textbf{7.13}                                                     \\
Relevance     & 5.93     & 7.30  & \textbf{9.56}                                                     & 3.34     & 5.82 & \textbf{7.32}                                                     \\
Specificity   & 5.04     & 7.01 & \textbf{9.06}                                                     & 5.07     & 6.25 & \textbf{7.44}                                                     \\
BLEU          & 0.31     & 0.68 & \textbf{0.95}                                                     & 0.12     & 0.28 & \textbf{0.36}                                                     \\
ROUGE-L       & 0.35     & 0.71 & \textbf{0.95}                                                     & 0.29     & 0.41 & \textbf{0.49}                                                     \\ \hline
\end{tabular}
\caption{Experimental results per method, averaged over single- and multi-hop free-text questions. Experiment done with questions generated on a selection of 8 industries (for local) and 8 industry pairs (for cross-industry) with a temperature of 0.5.}
\label{tab:qa_evals_methods_free}
\end{table}

Tables \ref{tab:qa_evals_methods} and \ref{tab:qa_evals_methods_free} show significant improvements in question quality across the three methods for both MCQs and free-text questions - there is a steady increase in scores for faithfulness, relevance, and consistency, as well as, for MCQs, the percentage of questions that have a single best answer. On average (across faithfulness, relevance, and specificity), local and cross-industry MCQs generated by the CoT + few-shot method are 57.6\% and 72.5\% better than the baseline method, respectively. For free-text questions, these respective improvements are 65.0\% and 68.7\%. \\

The CoT + few-shot method yields the best questions for both local and cross-industry spans, though cross-industry questions consistently score lower than local questions, not scoring above 8 on average. Cross-industry free-text questions also have much lower BLEU and ROUGE-L scores on average than local questions. Free-text questions score slightly lower than MCQs in general, and the extent of the difference in quality varies - free-text local questions, for example, score 8.75 on faithfulness, which is significantly less than the 9.45 scored by MCQs, though the difference between their relevance and specificity is much smaller.

\begin{table}[H]
\centering
\small
\begin{tabular}{lcccccccc}
\hline
\multirow{2}{*}{} & \multicolumn{4}{c}{MCQ}              & \multicolumn{4}{c}{Free-text} \\ \cline{2-9} 
                  & 0.0     & 0.2     & 0.5     & 1.0     & 0.0    & 0.2   & 0.5   & 1.0   \\ \cline{2-9} 
Faithfulness      & \textbf{8.83}    & 8.58    & 8.57    & 8.04    & 7.21   & 6.93  & \textbf{7.94}  & 7.63  \\
Relevance         & \textbf{9.10}    & 8.92    & 8.77    & 8.21    & 7.95   & 7.68  & \textbf{8.44}  & 8.38  \\
Specificity       & \textbf{8.50}    & 8.36    & 8.47    & \textbf{8.50}    & 8.08   & 8.13  & \textbf{8.25}  & 8.00  \\
BLEU              & --      & --      & --      & --      & 0.49   & 0.42  & \textbf{0.66}  & \textbf{0.66} \\
ROUGE-L           & --      & --      & --      & --      & 0.60   & 0.53  & 0.72  & \textbf{0.75}  \\
SBA (\%)          & 92.19\% & \textbf{92.97\%} & 92.19\% & 87.50\% & -      & -     & -     & -     \\ \hline
\end{tabular}
\caption{Temperature sensitivity for free-text and MCQ generated using the CoT + Few-shot method on a selection of 8 industries (for local questions) and 8 industry pairs (for cross-industry questions). Results averaged across single-/multi-hop and local/cross-industry questions. SBA (\%) stands for the percentage of questions that have a `single best answer'.}
\label{tab:temperature-comparison}
\end{table}

Table \ref{tab:temperature-comparison} shows that question quality varies little with LLM temperature, though small trends are observed. MCQs tend to score better with lower temperatures, though this is not the case for specificity and SBA percentages, which show no real trend. Free-text questions seem to score best with higher temperatures - faithfulness, relevance, and specificity are highest with a temperature of 0.5, while ROUGE-L is highest with a temperature of 1.

\subsection{Post-processing}

The efficacy of the quality improvement function is tested on the questions generated using the CoT + few-shot method. There are no standard accepted thresholds for faithfulness and relevance within LLM text generation literature. Generally, any score above 0.5 can be deemed to be acceptable, as more than half the generated text is truthful and related to the context. For the quality improvement function, we therefore base our threshold selection on the experimental results for CoT + few-shot shown in Tables \ref{tab:qa_evals_methods} and \ref{tab:qa_evals_methods_free}. We set a conditional threshold - local questions must score at least 9 on all metrics, while cross-industry questions must score at least 7 (both metrics are out of 10). \\

The post-processing functions are cost-intensive operations. In particular, the quality improvement function involves passing the entire question and source document(s) through the quality improvement function as well as re-evaluating the question on all metrics. Therefore, the application of the quality improvement and generalisation functions to the entire dataset is left as future work. These functions were tested qualitatively, with two examples shown below.
\newpage
\begin{lstlisting}[language=JSON,firstnumber=1,label={lst:quality_improvement},caption={Application of quality improvement post-processing on a cross-industry MCQ}]
Original question:
"question": "What are the metrics for Coal Operations and Apparel, Accessories & Footwear industries.",
    "optionA": "Production of thermal coal, Production of metallurgical coal, Number of Tier 1 suppliers and suppliers beyond Tier 1",
    "optionB": "Production of thermal coal, Number of Tier 1 suppliers and suppliers beyond Tier 1",
    "optionC": "Production of metallurgical coal, Number of Tier 1 suppliers and suppliers beyond Tier 1",
    "optionD": "Production of thermal coal, Production of metallurgical coal",
    "optionE": "Number of Tier 1 suppliers and suppliers beyond Tier 1",
"answer": "A"
Original specificity score: 5

Improved question:
"question": "List the activity metrics for Coal Operations and Apparel, Accessories & Footwear industries.",
    "optionA": "Production of thermal coal, Production of metallurgical coal, Number of Tier 1 suppliers and suppliers beyond Tier 1",
    "optionB": "Production of thermal coal, Number of Tier 1 suppliers and suppliers beyond Tier 1",
    "optionC": "Production of metallurgical coal, Number of Tier 1 suppliers and suppliers beyond Tier 1",
    "optionD": "Production of thermal coal, Production of metallurgical coal",
    "optionE": "Number of Tier 1 suppliers and suppliers beyond Tier 1",
"answer": "A"
New specificity score: 9
\end{lstlisting}

\begin{lstlisting}[language=JSON,firstnumber=1,label={lst:generalisation_example},caption={Application of generalisation post-processing on a cross-industry MCQ}]
"original question": "What is the code for the 'Gross global Scope 1 emissions, percentage covered under emissions-limiting regulations' metric in the Coal Operations industry?",
    "optionA": "EM-CO-110a.2",
    "optionB": "EM-CO-140a.1",
    "optionC": "EM-CO-110a.1",
    "optionD": "EM-CO-420a.3",
    "optionE": "EM-CO-000.A",
"answer": "C",

"refined question": "What is the code for the 'Gross global Scope 1 emissions, percentage covered under emissions-limiting regulations' metric in the fossil fuel operations industry?"
\end{lstlisting}

\subsection{Final Question Evaluation}

The final questions are generated using the CoT + few-shot method using all industries and industry pairs. 1,179 questions are generated using 68 industries (for local questions) and 44 industry pairs (for cross-industry questions) with question number per type shown in Table \ref{tab:final_qa_numbers}. The average quality results of these questions is shown in Tables \ref{tab:perindustry_results} and \ref{tab:per_hop}.

\begin{table}[H]
\centering
\begin{tabular}{lcccc}
\hline
           & \multicolumn{2}{c}{MCQ}                               & \multicolumn{2}{c}{Free-text}                         \\ \hline
           & \multicolumn{1}{l}{Local} & \multicolumn{1}{l}{Cross-industry} & \multicolumn{1}{l}{Local} & \multicolumn{1}{l}{Cross-industry} \\ \cline{2-5} 
Single-hop & 272                       & 93                        & 136                       & 88                        \\
Multi-hop  & 272                       & 94                        & 136                       & 88                        \\ \hline
\end{tabular}
\caption{Number of questions generated, per type, using the CoT + few-shot method.}
\label{tab:final_qa_numbers}
\end{table}

\begin{table}[H]
\centering
\begin{tabular}{lcccc}
\hline
              & \multicolumn{2}{c}{MCQ}  & \multicolumn{2}{c}{Free-text} \\ \cline{2-5} 
              & Local   & Cross-industry & Local     & Cross-industry    \\ \cline{2-5} 
No. Questions & 544     & 187            & 272       & 176               \\
Faithfulness  & \textbf{9.40}    & 7.64           & 8.32      & 6.42              \\
Relevance     & \textbf{9.68}    & 7.65           & 8.78      & 6.88              \\
Specificity   & \textbf{9.06}    & 7.15           & 8.80      & 8.22              \\
BLEU          & -       & -              & \textbf{0.93}      & 0.34              \\
ROUGE-L       & -       & -              & \textbf{0.95}      & 0.46              \\
SBA (\%)      & \textbf{92.47\%} & 91.45\%        & -         & -                 \\ \hline
\end{tabular}
\caption{Results per question span. Results are averaged across single- and multi-hop. Questions are generated using a temperature of 0.5. SBA (\%) stands for the percentage of questions that have a `single best answer'.}
\label{tab:perindustry_results}
\end{table}

\begin{table}[H]
\centering
\begin{tabular}{lcccc}
\hline
              & \multicolumn{2}{c}{MCQ} & \multicolumn{2}{c}{Free-text} \\ \cline{2-5} 
              & Single-hop  & Multi-hop & Single-hop     & Multi-hop    \\ \cline{2-5} 
No. Questions & 365         & 366       & 224            & 224          \\
Faithfulness  & \textbf{9.06}        & 7.98      & 7.52           & 7.23         \\
Relevance     & \textbf{8.84}        & 8.49      & 7.79           & 7.86         \\
Specificity   & 8.20         & 8.00         & \textbf{8.65}           & 8.35         \\
BLEU          & -           & -         & \textbf{0.76}           & 0.51         \\
ROUGE-L       & -           & -         & \textbf{0.82}           & 0.60          \\
SBA (\%)      & \textbf{93.43\%}     & 90.44\%   & -              & -            \\ \hline
\end{tabular}
\caption{Results per question complexity. Results are averaged across local and cross-industry. Questions are generated using a temperature of 0.5. SBA (\%) stands for the percentage of questions that have a `single best answer'.}
\label{tab:per_hop}
\end{table}

The average quality score (across faithfulness, relevance, and specificity) across all question types is 8.16. Local questions score 9 on average, while cross-industry ones score 7.32. Multiple-choice questions score 8.42 on average, while free-text questions score 7.9. Single-hop questions score 8.34 while multi-hop score 8 on average. \\

Based on these results, two general trends are observed: (1) `simpler' questions (i.e. local and single-hop) score better than more complex questions (i.e. cross-industry and multi-hop), and (2) MCQs score better than free-text questions. Among both MCQs and free-text questions, there is a bigger difference in quality between local and cross-industry questions than between single- and multi-hop questions. On average, single-hop MCQs score 6.7\% better than multi-hop MCQs, while local MCQs score 25.4\% better than cross-industry MCQs. For free-text questions, these differences are 2.2\% and 21.4\%, respectively.

\section{Discussion}

\subsection{Question Embeddings}

Figure \ref{fig:question_embeddings} shows two key trends - the question embeddings across different industry groups are generally distinct, while the embeddings within industry groups are similar. This means that, while it is easy to distinguish between the industry groups a query relates to, it is more difficult to distinguish between industries within the group based on the query.

\subsection{Experimental Results} \label{qa_experiments_discussion}

Each question generation method offered additional advantages for question quality. Table \ref{tab:qa_ablation} below displays the incremental improvements attributed to each method relative to the previous one. Note that the percentages for `Few-shot vs CoT' are calculated on a larger baseline than the percentages for `CoT vs baseline', so although the contributions of each additional method may seem similar, the absolute contributions of Few-shot are larger. 

\begin{table}[H]
\centering
\begin{tabular}{lcccc}
\hline
                     & \multicolumn{2}{c}{Local}                                                                                              & \multicolumn{2}{c}{Cross-industry}                                                                                    \\ \cline{2-5} 
                     & \begin{tabular}[c]{@{}c@{}}CoT \\ vs baseline\end{tabular} & \begin{tabular}[c]{@{}c@{}}Few-shot\\ vs CoT\end{tabular} & \begin{tabular}[c]{@{}c@{}}CoT\\ vs baseline\end{tabular} & \begin{tabular}[c]{@{}c@{}}Few-shot\\ vs CoT\end{tabular} \\ \cline{2-5} 
\multicolumn{1}{c}{} & \multicolumn{4}{c}{MCQ}                                                                                                                                                                                                                        \\ \cline{2-5} 
Faithfulness         & 17.50\%                                                    & 30.34\%                                                   & 8.60\%                                                    & 26.90\%                                                   \\
Relevance            & 24.48\%                                                    & 24.42\%                                                   & 71.39\%                                                   & 32.55\%                                                   \\
Specificity          & 33.16\%                                                    & 23.69\%                                                   & 25.20\%                                                   & 21.87\%                                                   \\
\textbf{Average}     & \textbf{25.05\%}                                           & \textbf{26.15\%}                                          & \textbf{35.06\%}                                          & \textbf{27.10\%}                                          \\ \cline{2-5} 
\multicolumn{1}{c}{} & \multicolumn{4}{c}{Free-text}                                                                                                                                                                                                                  \\ \cline{2-5} 
Faithfulness         & 25.35\%                                                    & 22.89\%                                                   & 15.72\%                                                   & 21.05\%                                                   \\
Relevance            & 23.10\%                                                    & 30.96\%                                                   & 74.25\%                                                   & 25.77\%                                                   \\
Specificity          & 39.09\%                                                    & 29.24\%                                                   & 23.27\%                                                   & 19.04\%                                                   \\
\textbf{Average}     & \textbf{29.18\%}                                           & \textbf{27.70\%}                                          & \textbf{37.75\%}                                          & \textbf{21.96\%}                                          \\ \hline
\end{tabular}
\caption{Percentage improvements in question quality across different methods. Calculated based on data in Tables \ref{tab:qa_evals_methods} and \ref{tab:qa_evals_methods_free}}
\label{tab:qa_ablation}
\end{table}

The percentage contributions of each additional method to question quality are mixed. Integrating CoT reasoning into the prompt yields average quality improvements in the range of 25\% to 37.8\%, with cross-industry questions benefiting most from this method. This suggests that guiding the LLM through a three-step process for creating questions helps it to navigate through multiple documents and isolate relevant chunks from each of them to form a question. \\

Few-shot prompting yields additional benefits ranging from 22\% to 27.7\%, and these benefits are mixed across local and cross-industry questions. However, in absolute terms, Tables \ref{tab:qa_evals_methods} and \ref{tab:qa_evals_methods_free} show that local questions are higher quality than cross-industry ones, scoring an average of 9.29 (across both MCQ and free-text) versus 7.52. A similar observation is made when comparing MCQs with free-text questions - though they both benefit equally from the advanced methods, MCQs are (albeit to a lesser extent) higher quality in absolute terms, scoring an average of 8.60 versus 8.21. \\

These observations suggest that there exists a limit to LLMs' capabilities of generating questions using prompt methods. It may struggle with using multiple documents to create questions, as well as ensuring that free-text questions are as accurate as multiple-choice ones. \\

In the following subsections, we discuss the quality issues encountered with each method, providing examples that have contributed to their respective quality scores.

\subsubsection{Baseline Prompting} 

Several issues are encountered in the baseline question generation method, as it does not guide the LLM on how to create questions and does not provide any example structures it should aim for. Four examples are shown below to highlight the biggest issues with the baseline approach (in addition to standard faithfulness problems).

\begin{lstlisting}[language=JSON,firstnumber=1,label={lst:memprompt},caption={Question Example 1: Local single-hop free-text}]
"question": "How many incidents of non-compliance with water quality permits, standards, and regulations have occurred for coal operations?",
"answer": "The entity shall disclose the total number of instances of non-compliance, including violations of a technology-based standard and exceedances of quantity and/or quality-based standards.",
\end{lstlisting}

\textit{Issue \#1: Questions are not related to sustainability reporting.} This is depicted in the example above, which displays phrasing that is not in line with user queries about how to report on the metrics. This in turn makes the answer, which is in fact correct based on the source, irrelevant to the question. A more relevant phrasing for this question would be ``How should I disclose incidents of non-compliance with water quality permits, standards, and regulations in the Coal Operations industry?". This issue is captured and measured by the relevance metric.

\begin{lstlisting}[language=JSON,firstnumber=1,label={lst:memprompt},caption={Question Example 2: Local single-hop MCQ}]
"question": "According to the topic summary, how can companies in this industry be better positioned to meet consumer demand and reduce future liabilities?",
"optionA": "By focusing solely on product innovation",
"optionB": "By reducing production costs at all costs",
"optionC": "By addressing environmental concerns and enhancing product take-back",
"optionD": "By increasing marketing budgets",
"optionE": "By outsourcing all manufacturing processes",
"answer": "C. By addressing environmental concerns and enhancing product take-back",
\end{lstlisting}

\textit{Issue \#2: Questions are vague, making the correct answer obvious.} In the example above, there is no specific reference to what piece of the document, and what industry, the question refers to. Nevertheless, the question is easily answerable even without knowing the context, as options A, B, D, and E are very obviously wrong. This issue is measured by the domain specificity metric, which shows the lowest scores of all metrics for the baseline method, indicating that this is a particular issue with this method.

\begin{lstlisting}[language=JSON,firstnumber=1,label={lst:memprompt},caption={Question Example 3: Cross-industry single-hop MCQ}]
"question": "Which of the following is a key strategy home builders can use to mitigate environmental legal risks?",
"optionA": "Ensuring all homes are certified to a third-party green building standard",
"optionB": "Implementing corrective actions in response to any legal proceedings related to environmental regulations",
"optionC": "Focusing development only in regions with low baseline water stress",
"optionD": "Both B and C",
"optionE": "All of the above",
"answer": "B",
"industries": ["b8-construction-materials","b35-home-builders"]
\end{lstlisting}

\textit{Issue \#3: Cross-industry questions do not cover all required industries.} In the example above, not all the required industries are covered by the question. This issue is captured and measured by the relevance metric, contributing to its low score.

\begin{lstlisting}[language=JSON,firstnumber=1,label={lst:memprompt},caption={Question Example 4: Cross-industry single-hop MCQ}]
"question": "Which of the following is a key risk for home builders related to climate change adaptation?",
    "optionA": "Increased costs associated with flood insurance for homes in 100-year flood zones",
    "optionB": "Difficulty obtaining permits for developments in water-stressed regions",
    "optionC": "Reduced long-term demand for homes in volatile climate regions",
    "optionD": "Need for more robust construction materials to withstand extreme weather events",
    "optionE": "Higher energy costs for cooling homes in regions experiencing rising temperatures",
"answer": "A"
"true correct answers": A, B, C
\end{lstlisting}

\textit{Issue \#4: Questions do not have a single best answer.} The example question above has three answer options that are correct, though the LLM has selected only one. Multiple-choice questions often have multiple correct answers listed in the options. This is driven by questions being vague, inviting multiple ways of answering them. The LLM simply chooses one of its generated answer options to be the `correct' one, however several of the other given options can also be deemed correct based on the source information. Having multiple correct answer options artificially deflates the accuracy of RAG systems tested on the MCQs. The lack of a single best answer affects one in three questions generated by the baseline method (see Table \ref{tab:qa_evals_methods}).

\subsubsection{CoT Prompting}

The CoT approach addresses some of the issues encountered in the baseline approach, and the highest quality improvements are seen for cross-industry questions, where the quality improved by an average of 36.3\% across all metrics (the average improvement for local questions is 27\%). Most notably, cross-industry questions become 72.8\% more relevant and 24.2\% more specific. Nevertheless, in absolute terms, these questions do not achieve a satisfactory level, with scores being only slightly above average. Some sample questions are displayed below to highlight persistent issues with question quality.

\begin{lstlisting}[language=JSON,firstnumber=1,label={lst:cot_example1},caption={Question Example 1: Local single-hop free-text}]
"question": "What is the percentage of the company's Scope 1 emissions that are covered under emissions-limiting regulations?",

"answer": "The percentage shall be calculated as the total amount of gross global Scope 1 GHG emissions (CO2-e) that are covered under emissions-limiting regulations divided by the total amount of gross global Scope 1 GHG emissions (CO2-e).",
\end{lstlisting}

\textit{Issue \#1: Questions are vague.} This question has no reference to what type of industry it is referring to and is therefore applicable to all industries. It would be nearly impossible for any RAG system to get this question correct with no further context.

\begin{lstlisting}[language=JSON,firstnumber=1,label={lst:cot_example2},caption={Question Example 2: Cross-industry single-hop MCQ}]
"question": "Which industry discusses the percentage of products by revenue that contain IEC 62474 declarable substances?",
    "optionA": "Metals & Mining",
    "optionB": "Electrical & Electronic Equipment",
    "optionC": "Both industries",
    "optionD": "Neither industry",
    "optionE": "Not specified",
    "answer": "B"
    "industries": ["b10-metals-and-mining","b49-electrical-and-electronic-equipment"]
\end{lstlisting}

\textit{Issues \#2 and \#3: Questions do not cover all industries required, and are not representative of real-life user queries  for sustainability reporting.} This question is supposed to be comparative between two different industries. Instead, it lists industries as multiple-choice options, a format that is not user-typical. The question itself doesn't refer to both industries it is given. These issues contribute to lower relevancy and specificity.

\subsubsection{CoT + Few-shot Prompting}

This is the final selected method for question generation, which displays the best quality metric scores and the highest percentage of questions that pass the SBA test out of all the methods. These improvements can be attributed to the precise wording and structuring of the few-shot examples provided to the LLM, as described in Section \ref{qa_design}. \\

Below are some sample questions that depict high-quality questions that are representative of real-life user queries.

\begin{lstlisting}[language=JSON,firstnumber=1,label={lst:few_example2},caption={Question Example 1: Local multi-hop MCQ}]
{
"question": "What is the category of the metric 'Total landfill gas generated' in the Waste Management industry and how should this metric be calculated?",
      "optionA": "Quantitative; calculated in millions of British Thermal Units (MMBtu)",
      "optionB": "Discussion and Analysis; calculated based on engineering estimates",
      "optionC": "Quantitative; calculated in metric tons (t)",
      "optionD": "Qualitative; based on expert interviews",
      "optionE": "Quantitative; calculated in gigajoules (GJ)",
"answer": "A",
"reference_text": [
    "1 The entity shall disclose (1) the total amount, in millions of British Thermal Units (MMBtu), of landfill gas generated from its owned or operated facilities.",
    "1.1 Landfill gas is defined as gas produced as a result of anaerobic decomposition of waste materials in the landfill."
      ],
"pages": ["401"]
\end{lstlisting}

\begin{lstlisting}[language=JSON,firstnumber=1,label={lst:few_example1},caption={Question Example 2: Cross-industry multi-hop MCQ}]
"question": "Give me the codes for all the 'Quantitative' metrics in Table 1 for Insurance and Real Estate Services industries.",
      "optionA": "FN-IN-410b.1, FN-IN-450a.1, FN-IN-450a.2, IF-RS-410a.1, IF-RS-410a.2, IF-RS-410a.3",
      "optionB": "FN-IN-410b.1, FN-IN-450a.1, FN-IN-450a.2, IF-RS-410a.1",
      "optionC": "FN-IN-410b.1, FN-IN-450a.1, IF-RS-410a.1, IF-RS-410a.2",
      "optionD": "FN-IN-450a.1, FN-IN-450a.2, IF-RS-410a.1, IF-RS-410a.3",
      "optionE": "FN-IN-410b.1, FN-IN-450a.1, IF-RS-410a.2, IF-RS-410a.3",
"answer": "A",
"reference_text": [
    "| Policies Designed to Incentivize Responsible Behavior | Net premiums written related to energy efficiency and low carbon technology | Quantitative | Reporting currency | FN-IN-410b.1 |",
    "| Environmental Physical Risk Exposure | Probable Maximum Loss (PML) of insured products from weather-related natural catastrophes 21 | Quantitative | Reporting currency | FN-IN-450a.1 |",
    "| Environmental Physical Risk Exposure | Total amount of monetary losses attributable to insurance payouts from (1) modeled natural catastrophes and (2) non-modeled natural catastrophes, by type of event and geographic segment (net and gross of reinsurance) 22 | Quantitative | Reporting currency | FN-IN-450a.2 |",
    "| Sustainability Services | Revenue from energy and sustainability services 55 | Quantitative | Reporting currency | IF-RS-410a.1 |",
    "| Sustainability Services | (1) Floor area and (2) number of buildings under management provided with energy and sustainability services | Quantitative | Square feet (ft\u00b2), Number | IF-RS-410a.2 |",
    "| Sustainability Services | (1) Floor area and (2) number of buildings under management that obtained an energy rating | Quantitative | Square feet (ft\u00b2), Number | IF-RS-410a.3 |"],
"pages": ["Page 156","Page 391"],
\end{lstlisting}

Nevertheless, these questions are not perfect, and there are limits to what types of questions can be generated using advanced prompting methods. One such limit was encountered when trying to generate questions spanning more than two industries, where the LLM was not able to capture information from all industries to create a question. Instead, it selected one or two of the given industries and made a question based on that. This is not a significant problem for our application, as such complicated questions may be less likely to occur in our domain, but presents some room for potential future work on question-generation methods.

\subsubsection{Temperature}

Table \ref{tab:temperature-comparison} shows the evaluation of questions generated using the CoT + few-shot method with different LLM temperature settings. The differences in evaluation scores across temperature values are small, showing that LLM temperature does not have a strong impact on question quality. This is because the LLM is guided to follow certain question styles through few-shot prompting, removing much scope for LLM creativity (which is what temperature dictates). Given the small variance, and to ensure all questions are created with consistent settings, the decision is made to use a temperature of 0.5 for all question generation. 

\subsection{Final Question Evaluation}

The results in Tables \ref{tab:perindustry_results} and \ref{tab:per_hop} show that the CoT + few-shot generation method produces high-quality questions across all formats. Nevertheless, table \ref{tab:final_ablation} shows that question span is still a significant driver of question quality, as is question complexity, although to a much lesser extent. This highlights that LLMs possess some ability to reason in multiple steps but struggle more with utilising data from multiple source documents. A manual evaluation of a subset of the questions show that they are high quality (as depicted in the examples shown in Section \ref{qa_experiments_discussion}), though conducting human evaluation on the entire dataset is left as future work.

\begin{table}[H]
\centering
\begin{tabular}{lcc}
\hline
             & Question Span           & Question Complexity     \\ \hline
             & Local vs Cross-industry & Single-hop vs Multi-hop \\ \cline{2-3} 
             & \multicolumn{2}{c}{MCQ}                           \\ \cline{2-3} 
Faithfulness & 23.04\%                 & 13.53\%                 \\
Relevance    & 26.54\%                 & 4.12\%                  \\
Specificity  & 26.71\%                 & 2.50\%                  \\
\textbf{Average}      & \textbf{25.43\% }                &\textbf{ 6.72\%}                  \\ \cline{2-3} 
             & \multicolumn{2}{c}{Free-text}                     \\ \cline{2-3} 
Faithfulness & 29.60\%                 & 4.01\%                  \\
Relevance    & 27.62\%                 & -0.89\%                 \\
Specificity  & 7.06\%                  & 3.59\%                  \\
\textbf{Average}      & \textbf{21.42\% }                & \textbf{2.24\%}                  \\ \hline
\end{tabular}
\caption{Percentage differences in question quality of simpler questions relative to more difficult questions, based on question span and question complexity. Calculated based on data in Tables \ref{tab:perindustry_results} and \ref{tab:per_hop}.}
\label{tab:final_ablation}
\end{table}

For the final dataset, 116 questions are removed as they do not have a single best answer, leaving 1,063 questions to be used for evaluating RAG systems.

\chapter{Domain-Specific Assistant}

This chapter presents the techniques implemented to design a useful domain-specific knowledge assistant. We start by defining the knowledge scope of the chatbot. Then, we describe the methods investigated to design such a chatbot, including RAG, fine-tuning, and classification techniques, and propose two novel architectures for the chatbot - a RAG-based pipeline and an LLM-based pipeline. The dataset developed in Chapter 3 is used along with the IFRS markdown texts to train the classification models, fine-tune LLMs, and conduct experiments to evaluate all techniques. We present the results of the experiments as well as the final performance of the proposed architectures on the QA dataset.

\section{Knowledge Scope}

The knowledge scope for the assistant is defined by the IFRS Sustainability Reporting standards detailed in Section \ref{data_collection}. To be `useful' for answering user queries based on this knowledge, the assistant must be able to retrieve, comprehend, and reason over this data to answer questions, while being constrained from answering out-of-scope queries.

\section{Methods} 

First, an initial investigation is conducted into different RAG techniques by conducting experiments using the QA dataset. Then, an LLM is fine-tuned and tested on the QA dataset. Next, several methods are explored to design an industry classifier, including training an NLP model, two machine learning (ML) methods, a multi-layer perceptron (MLP), and a prompt-based LLM classifier. Finally, two chatbot architectures are proposed - a RAG pipeline and an LLM-based pipeline.

\subsection{Retrieval Augmented Generation (RAG)}

This section explores a selection of RAG techniques for the domain-specific assistant. The steps of a basic RAG pipeline (shown in Figure \ref{fig:basic_rag} of Chapter 2) are outlined below, alongside the alternative methods tested for each step, which are described in detail in the following sections.

\begin{enumerate}
    \item \textbf{Indexing:} text is extracted from the documents and broken down into smaller chunks, which are embedded into vector representation using an embedding model and stored in a vector DB. \\
    \textit{Chunking methods tested:} fixed-size chunks of 256, 512, and 1024 tokens, page, rolling window, semantic chunking, and custom markdown chunking.
    \item \textbf{Retrieval:} When a user asks a question, the query is embedded (using the same model as the data) into a vector and is compared to the stored vectors to find the most similar matches. In the basic case, the query is embedded as is (with no transforms applied), and KNN is used as the retrieval technique.\\
    \textit{Retrieval techniques tested:} KNN and hybrid.\\
    \textit{Query Transformation techniques tested:} HyDE, multi-query.
    \item \textbf{Generation: }This is the final step, where the user's question and the best-matching information are combined and given to an LLM to create an answer.
    \textit{LLMs tested:} Google Gemma 2B, Meta’s Llama 2 13B, Llama 3.1 8B, Llama 3 70B, Mixtral 8x7B, and fine-tuned Llama 3.1 8B.
\end{enumerate}

\subsubsection{Models}

To assess the impact of LLM size and capabilities on RAG performance, a selection of models is tested. The models included are smaller parameter models like Google's Gemma 2B and Meta's Llama 3.1 8B, mid-sized models such as Llama 2 13B, and larger models like Llama 3 70B and Mixtral 8x7B. The models differ in their training data, architecture, and parameter count, which can influence their reasoning capabilities. For instance, the impact of model scale on performance can be examined using the Llama family models (2 13B, 3.1 8B, and 3 70B) share a similar architecture but vary in size. Mixtral 8x7B, on the other hand, utilises a mixture-of-experts architecture \cite{shazeer2017outrageouslylargeneuralnetworks} and gives insight into how this compares to the other models. \\

The fine-tuned Llama 3.1 8B model described in Section \ref{finetuning} is also tested to evaluate the impact of domain-specific fine-tuning on performance.

\subsubsection{Data Chunking} \label{data_chunking_methods}

One limitation of LLMs is their restricted context window which limits their comprehension abilities for extensive documents \cite{liu2024lost, lmsys}. To prepare the markdown data extracted from the PDF reports for retrieval, it must be broken down into chunks - smaller, more focused segments that the LLM can process with greater precision. Data chunking can impact the performance of RAG systems by directly impacting the quality and relevance of retrieved context being inputted into the LLM for answer generation \cite{liu2024lost}. This project explores seven chunking strategies to determine the optimal approach for the Sustainability Reporting domain. The chunking strategies explored are fixed-size chunks of 256, 512, and 1024 tokens, page-based chunking, a rolling window approach, semantic chunking, and custom markdown chunking. \\

Fixed-size chunks of 512 tokens are a common baseline in RAG systems, balancing granularity with computational efficiency. To investigate the impact of chunk size on performance, 256- and 1024-token chunks are also evaluated. These larger chunks may provide less/more comprehensive context for complex queries but could potentially reduce retrieval precision by missing information or confusing the LLM.\\

One issue with fixed-length chunking strategies is that relevant information might be split across chunk boundaries. To address this, a rolling window approach with 512-token chunks and 10\% overlap is implemented to improve retrieval robustness by ensuring that key information is not inadvertently separated. \\

A page-based chunking strategy is implemented to preserve the original document structure and may be particularly helpful for queries referencing specific pages. Semantic chunking is also explored to create chunks based on topical coherence rather than arbitrary token counts. This method aims to preserve the semantic integrity of the content, potentially improving the relevance of retrieved chunks for complex queries. \\

Finally, this project leverages the markdown structure of the reports to implement a custom chunking strategy. Markdown chunking splits the reports by their headings at a chosen level of hierarchy (e.g. at the section/subsection level). Additionally, it chunks tables as a whole, separately from free-text content. Maintaining the integrity of tables reduces the risk of the retriever missing or misinterpreting relevant rows or columns of data. This method is akin to that introduced by Yepes et al \cite{yepes2024financial} for the finance domain. \\

The chunked data is embedded using the OpenAI embedding model text-embedding-3-small and stored in Pinecone, a vector database. Metadata is appended to each chunk containing the page number the text is taken from, the report name, the industry, and the type of information (free text or table) contained in the chunk. Chunking strategies are systematically evaluated on the diverse synthetic QA dataset to determine the most effective method for this domain. Results are shown in Section \ref{chunking_results}.

\subsubsection{Retrieval Techniques}

Retrieval techniques define how the most relevant context chunks for a query are determined, and choice of technique can significantly influence the effectiveness of RAG systems \cite{gao2024retrievalaugmented}. This study examines three widely used retrieval techniques: KNN and hybrid.\\

K-Nearest Neighbors (KNN) retrieves the K most similar chunks based on embedding cosine similarity. This is a simple and efficient method but may struggle with capturing complex relationships between queries and documents. Hybrid approaches combine embedding-based similarity with traditional text search techniques such as BM25 (Best Match), a search algorithm that selects documents according to their query relevance, using Term Frequency (TF), Inverse Document Frequency (IDF), and Document Length to compute a relevance score \cite{sawarkar2024blended}. This can be particularly effective for queries that contain both conceptual elements and key terms or phrases. These retrieval techniques are systematically evaluated to identify the most effective method for the sustainability reporting domain.  \\

Three further popular retrievers - SVM, linear regression, and Maximum Marginal Relevance (MMR) - are also tested, with results shown in Appendix \ref{retriever_appx}.

\subsubsection{Query Transforms}

The study explores two query transformation techniques: Hypothetical Document Embeddings (HyDE), and multi-query expansion. Each method is an approach to modifying or expanding the original query to potentially improve retrieval performance. The baseline approach involves using the original query without any modifications. \\

Hypothetical Document Embeddings (HyDE) \cite{gao2022precise} is a query transformation technique that leverages LLMs to generate a hypothetical document that could answer the query, and then uses the embedding of this document for retrieval. The process involves inputting the original query to an LLM, which then generates a hypothetical document that could potentially answer the query. This hypothetical document is then embedded and used, instead of the original query, to retrieve real documents. HyDE aims to bridge the semantic gap between queries and documents by creating a more comprehensive representation of the user's information needs. Gao et al. \cite{gao2022precise} demonstrated that HyDE can improve retrieval performance across various domains, particularly for complex or ambiguous queries. However, the hypothetical document is fake and therefore may contain false details, introducing hallucinations into the retrieval process. \\

Multi-query expansion \cite{medium_rag_not_working} involves generating multiple variations or reformulations of the original query to capture different aspects of the information needed. This technique analyses the original query to identify key terms or concepts and formulates alternative formulations of the query, often using synonyms, related terms, or different phrasings. These multiple queries are then used for retrieval using parallel vector searches and intelligent reranking to aggregate results. The goal of multi-query expansion is to improve recall by addressing vocabulary mismatch problems and capturing a broader range of relevant documents.

\subsection{Fine-tuning}\label{finetuning}

Another method for integrating domain-specific data into an LLM is fine-tuning (as discussed in Chapter 2). The Llama 3.1 8B model was fine-tuned using the final MCQ QA pairs to enhance its performance on multiple-choice questions without RAG, by learning the specific information and thus reducing the hallucination rate. LoRA finetuning was implemented using Together.AI. The performance of the fine-tuned Llama 3.1 8B was compared to both the original Llama 3.1 8B and a larger Llama 3.1 40B model on local and cross-industry questions.

\subsection{Industry Classification} 
Six models were implemented for multi-label industry classification: a natural language processing (NLP) approach, a machine learning (ML) approach, a Multi-Layer Perceptron (MLP) model, and two prompting-based approaches using GPT-4o and GPT-4o Mini. Each model was designed to output multiple industry labels per input question, accommodating up to 5 industries per question.

\subsubsection{NLP Model (DistilBERT)} 

The NLP model is based on DistilBERT \cite{sanh2020distilbertdistilledversionbert}, a smaller and more efficient version of BERT tailored for sequence classification. The model includes a pre-classifier layer and a final classifier with sigmoid activation, allowing for multi-label classification by outputting probabilities for each industry. Input text was tokenized using the distilbert-base-uncased tokenizer.

\subsubsection{ML Models (Random Forest and XGBoost)} 

The machine learning approach involved two models: a Random Forest classifier and an XGBoost model. Both models were trained on text embeddings generated using OpenAI's text-embedding-3-small model. The Random Forest model was composed of 200 decision trees, while the XGBoost model used 100 estimators. Each model was tuned to predict multiple industry labels per input, with a custom threshold applied to the output probabilities to determine the final industry predictions.

\subsubsection{MLP Classifier}

The Multi-Layer Perceptron (MLP) classifier was structured with two hidden layers and trained using the same embeddings as the ML models, using a sigmoid activation in the output layer to generate probabilities for each industry, allowing for multi-label classification. This model was selected to learn non-linear patterns in the input data.

\subsubsection{LLM Classifier} \label{llm_classifier}

The prompting-based approach used GPT-4o and GPT-4o Mini to classify industries using a custom prompt containing industry descriptions. The returned predictions were structured using a Pydantic model, allowing for variable industry outputs depending on the input query.

\subsubsection{Evaluation Metrics} 

The performance of each model was evaluated using the following metrics: \begin{itemize} \item Macro F1 Score: This metric calculates the harmonic mean of precision and recall for each industry label, treating each label equally, regardless of its frequency in the dataset. It provides a balanced measure of performance across all industry categories. \item Precision: Precision measures the proportion of correctly predicted industry labels out of all labels predicted by the model, indicating how accurate the model is when it makes a positive prediction. \item Recall: Recall calculates the proportion of correctly predicted industry labels out of all true industry labels, evaluating how well the model identifies all relevant industries. \item Hamming Loss: This metric quantifies the fraction of labels that are incorrectly predicted. It accounts for both false positives and false negatives, providing insight into how often the model makes incorrect label assignments. \end{itemize}

\subsection{Proposed Architecture}

For the final question-answering system, two architectures are proposed. The first one leverages the RAG techniques studied along with the GPT-4o Mini industry classifier. The second one uses LLMs throughout the entire pipeline. Both pipelines integrate a pre-processing step to stop the chatbot from answering queries that are not related to IFRS sustainability reporting standards. This is done by passing the user query through an LLM call to verify that it is relevant before proceeding. 

\subsubsection{Custom RAG Pipeline}

\begin{figure}
    \centering
    \includegraphics[width=0.7\linewidth]{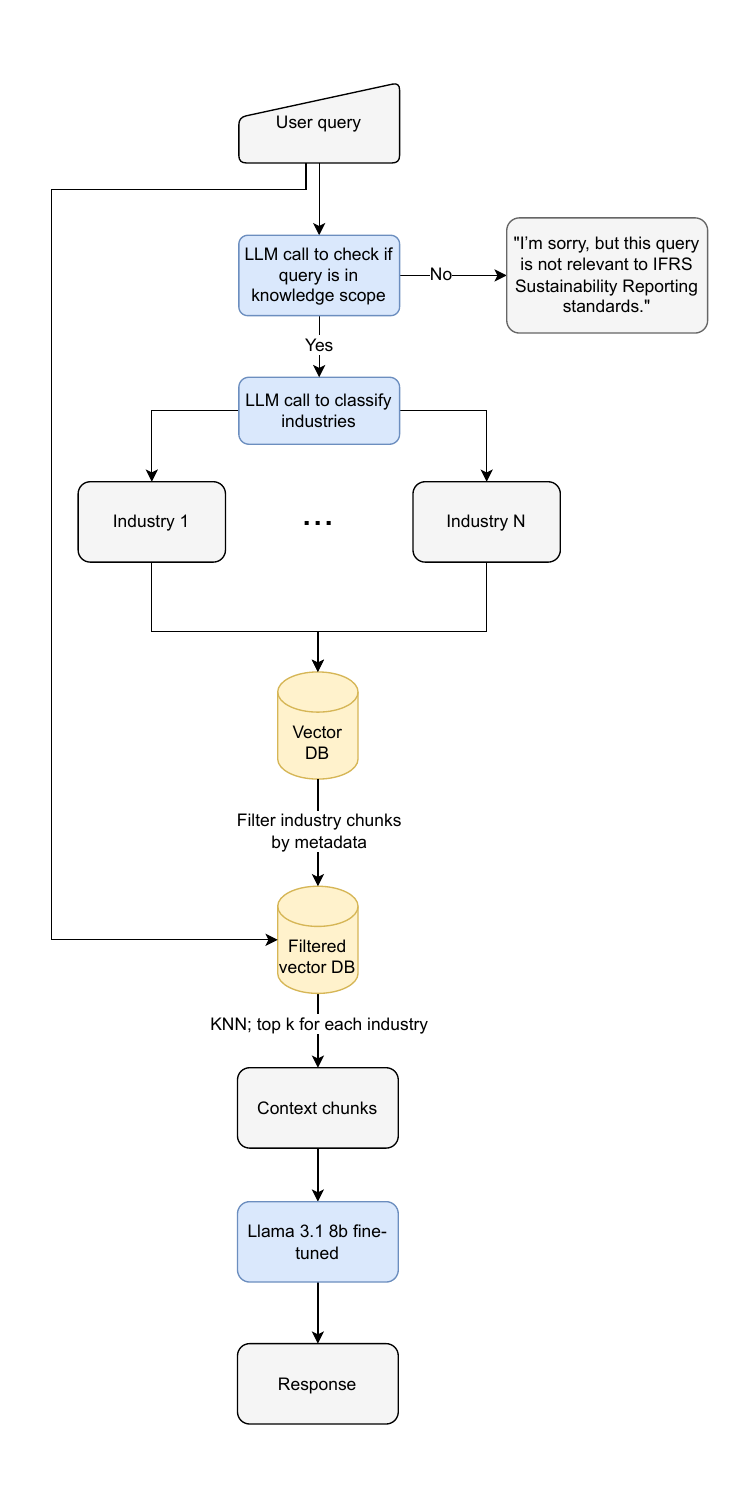}
    \caption{Custom RAG pipeline.}
    \label{fig:rag_pipeline}
\end{figure}

A custom RAG pipeline is designed that combines the strengths of the RAG methods tested. In particular, it integrates two novel elements - the prompt-based industry classifier and the fine-tuned Llama 3.1 8B model - with established RAG techniques to create a more targeted system for answering queries related to sustainability reporting standards. The pipeline is shown in Figure \ref{fig:rag_pipeline}.\\

User queries are checked for domain relevancy and passed into the LLM industry classifier described in Section \ref{llm_classifier}, which identifies the most likely industries the query is referring to. The chunks stored in the vector database, which were chunked by the custom markdown method, are then filtered using their metadata (since each chunk is labelled by industry as described in Section \ref{data_chunking_methods}) to constrain the database to only chunks from those industries. The top 5 most relevant chunks are then retrieved from the filtered database using KNN. These are passed to the fine-tuned Llama 3.1 8B model which uses them to generate a response.

\subsubsection{Fully LLM-based Pipeline}

\begin{figure}[H]
    \centering
    \includegraphics[width=0.75\linewidth]{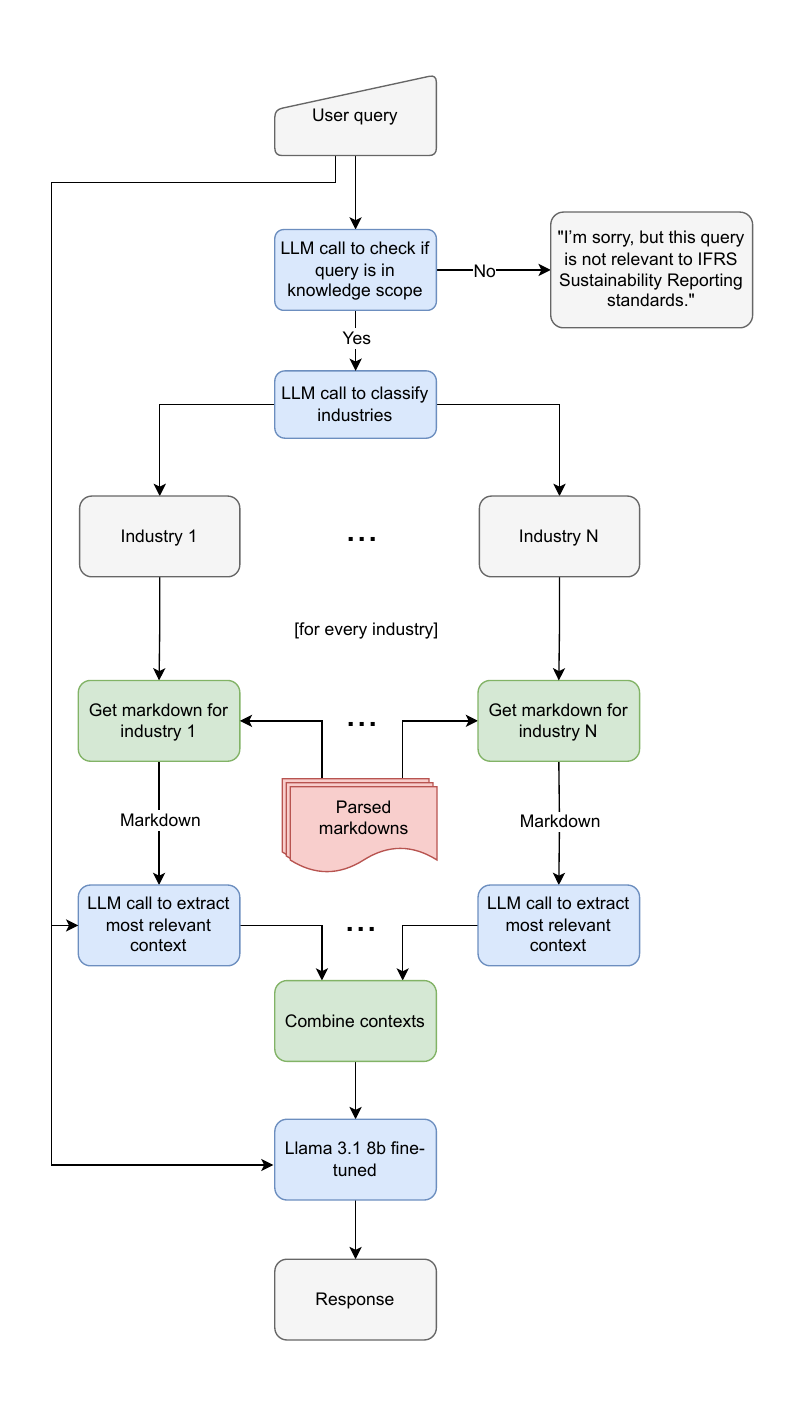}
    \caption{Custom LLM pipeline.}
    \label{fig:llm_rag}
\end{figure}

The second question-answering pipeline, shown in Figure \ref{fig:llm_rag}, is designed using LLMs throughout. After the query has been deemed relevant to the domain, the information retrieval and answer generation steps are as follows:
\begin{enumerate}
    \item The LLM classifier is used to output the most likely industries related to the query.
    \item The markdown report is selected for each industry. 
    \item For each industry separately, the associated industry markdown report is passed along with the user query to an LLM, which retrieves the chunks of context that are best suited to answer the question. The LLM used for this step is GPT-4o Mini as it has a long enough context length.
    \item The chunks for each industry are labelled with the industry, and all industry chunks are combined.
    \item The combined chunks are passed along with the user query to the fine-tuned Llama 3.1 8B model to generate an answer.
\end{enumerate}

This architecture displays two key differences relative to the RAG pipeline. Firstly, the system operates by passing inputs directly into LLMs, without embedding data or the query. This bypasses the need for similarity search methods for retrieving relevant chunks and instead leverages the reasoning abilities of the LLM to gauge relevance. Secondly, the data is not broken down into chunks beforehand. Instead, an LLM call takes as input the entire industry markdown to select the most relevant chunks based on the user query. This provides flexibility to tailor the number and content of the retrieved chunks to each query, rather than relying on a fixed chunking strategy and retrieval number. \\

\section{Results}

All methods are evaluated on both MCQ and free-text questions. Accuracy is used on MCQs, and BLEU and ROUGE-L are used for evaluation on free-text questions.

\subsection{RAG}

\subsubsection{Models}

\begin{table}[H]
\centering
\begin{tabular}{lcccccc}
\hline
\multirow{3}{*}{Model} & \multicolumn{2}{c}{MCQ} & \multicolumn{4}{c}{Free text} \\ \cline{2-7} 
 & Local & Cross-Industry & \multicolumn{2}{c}{Local} & \multicolumn{2}{c}{Cross-Industry} \\  
 & Accuracy & Accuracy & BLEU & ROUGE & BLEU & ROUGE \\ \hline
Google Gemma 2B         & 12.90 & 17.64 & 0.18 & 0.36 & 0.05 & 0.14 \\
Llama 2 13B             & 38.71 & 17.64 & 0.17 & 0.34 & 0.01 & 0.03 \\
Llama 3.1 8B            & 80.65 & 52.94 & 0.41 & 0.60 & 0.06 & 0.23 \\
\begin{tabular}[c]{@{}l@{}}Llama 3.1 8B \\ Fine-tuned\end{tabular} & \textbf{83.87} & 53.16 & \textbf{0.49} & \textbf{0.65} & \textbf{0.07} & \textbf{0.24} \\
Llama 3 70B             & 80.65 &\textbf{ 60.72} & 0.42 & 0.62 & 0.04 & 0.19 \\
Mixtral 8x7B            & 80.65 & 37.25 & 0.18 & 0.38 & 0.05 & 0.02 \\ \hline
\end{tabular}
\caption{RAG results with different LLM and baseline settings: 512-token chunking, KNN retrieval with top K=5 chunks, no query transform, and LLM temperature of 0.5.}
\label{tab:models}
\end{table}

Table \ref{tab:models} shows that Llama 3.1 models generally outperform others on both local and cross-industry questions, and little variance is observed across model size between Llama 3.1 8B and 40B. However, a large gap exists between Llama 3.1 8B and Llama 2 13B, which is a much less capable model despite being larger. Llama 2's performance is more comparable to Google Gemma 2B, showing the same performance in MCQ accuracy (17.64\%). \\

The fine-tuned Llama 3.1 8B model shows the best performance across all metrics for local questions, as measured by both MCQ accuracy and free-text BLEU and ROUGE. In particular, it shows better performance than the significantly larger model Llama 3 70B. However, the fine-tuned model's outperformance over the standard Llama 3.1 8B model is smaller for cross-industry questions, performing worse on these than the larger 70B model. This suggests that fine-tuning (when coupled with RAG) improves the model's ability to handle highly specific `factual' queries, though provides limited benefit for more complex queries, where larger models can leverage their enhanced reasoning power.\\

All models show a noticeable drop in performance for cross-industry questions compared to local ones. Nevertheless, one notable observation is that, while cross-industry performance is low, it is still above average since a random answer selection strategy would result in a 20\% accuracy for MCQs with 5 answer options.\\

Finally, it is of note that BLEU and ROUGE are generally very low, particularly for cross-industry questions. This is discussed in Section \ref{models_discussion}.

\subsubsection{Data Chunking}\label{chunking_results}

\begin{table}[H]
\centering
\begin{tabular}{lcccccc}
\hline
\multirow{3}{*}{Chunking} & \multicolumn{2}{c}{MCQ} & \multicolumn{4}{c}{Free text} \\ \cline{2-7} 
 & Local & Cross-Industry & \multicolumn{2}{c}{Local} & \multicolumn{2}{c}{Cross-Industry} \\   
 & Accuracy & Accuracy & BLEU & ROUGE & BLEU & ROUGE \\ \hline
Sentence 256    & 80.65 & 41.26 & 0.25 & 0.48 & \textbf{0.09} & \textbf{0.24} \\
Sentence 512    & 80.65 & 52.94 & \textbf{0.41} & \textbf{0.60} & 0.06 & 0.23 \\
Sentence 1024   & 38.71 & 31.09 & 0.28 & 0.49 & 0.06 & 0.22 \\
Sentence window & 81.32 & 54.55 & 0.04 & 0.19 & 0.05 & 0.15 \\
Page            & 65.12 & 37.45 & 0.03 & 0.16 & 0.06 & 0.17 \\
Semantic        & 83.30 & 63.90 & 0.33 & 0.55 & 0.06 & 0.23 \\
Custom Markdown & \textbf{84.52} & \textbf{69.10} & 0.04 & 0.19 & 0.04 & 0.15 \\ \hline
\end{tabular}
\caption{RAG results with different chunking methods and baseline settings: Llama 3.1 8B, KNN retrieval with top K=5 chunks, no query transform, and LLM temperature of 0.5.}
\label{tab:chunking}
\end{table}

Table \ref{tab:chunking} shows the performance of the Llama 3.1 8B model across different chunking strategies. For MCQ accuracy, the custom markdown chunking method performs best, achieving 84.52\% for local and 69.10\% for cross-industry questions. This is followed closely by semantic chunking at 83.30\% and 63.90\% respectively.\\

In the free text evaluation, sentence-based chunking strategies with 256 and 512 tokens generally outperform other methods, particularly for local scenarios. Sentence 512 chunking achieves the highest BLEU (0.41) and ROUGE (0.60) scores for local free-text responses. Performance drops significantly for cross-industry tasks across all chunking methods. The page-level chunking shows the poorest performance overall, particularly for free text tasks. \\

A notable result is that Sentence 1024 shows very poor performance for MCQs, but reasonable results on free-text questions. Reasons for this are discussed in Section \ref{chunking_discussion}.

\subsubsection{Retrieval Techniques}
\begin{table}[H]
\centering
\begin{tabular}{lcccccc}
\hline
\multirow{3}{*}{Retriever} & \multicolumn{2}{c}{MCQ} & \multicolumn{4}{c}{Free text} \\ \cline{2-7} 
 & Local & Cross-Industry & \multicolumn{2}{c}{Local} & \multicolumn{2}{c}{Cross-Industry} \\  
 & Accuracy & Accuracy & BLEU & ROUGE & BLEU & ROUGE \\ \hline
KNN     & \textbf{80.65} & \textbf{52.94} & \textbf{0.41} & \textbf{0.60} & \textbf{0.06} & \textbf{0.23} \\
Hybrid            & \textbf{80.65} & 47.05 & 0.37 & 0.58 & \textbf{0.06} & \textbf{0.23} \\ \hline
\end{tabular}
\caption{RAG results with different chunking methods and baseline settings: Llama 3.1 8B, retrieval of top K=5 chunks, no query transform, and LLM temperature of 0.5.}
\label{tab:retriever-comparison}
\end{table}

Table \ref{tab:retriever-comparison} presents the performance of the two retrieval techniques tested. The retrievers perform equally well on local MCQs, achieving 80.65\% accuracy, though KNN performs better on Cross-industry MCQs, achieving 52.94\% versus 47.05\%. KNN also displays the strongest performance on free-text questions, obtaining BLEU and ROUGE scores of 0.41 and 0.60 for local free text responses. Both retrievers exhibit a significant drop in performance when moving from local to cross-industry tasks.

\subsubsection{Query Transforms}

\begin{table}[H]
\centering
\begin{tabular}{lcccccc}
\hline
\multirow{3}{*}{Query Transform} & \multicolumn{2}{c}{MCQ} & \multicolumn{4}{c}{Free text} \\ \cline{2-7} 
 & Local & Cross-Industry & \multicolumn{2}{c}{Local} & \multicolumn{2}{c}{Cross-Industry} \\  
 & Accuracy & Accuracy & BLEU & ROUGE & BLEU & ROUGE \\ \hline
None        & 80.65 & 52.94 & \textbf{0.41} & \textbf{0.60} & 0.06 & 0.23 \\
HyDE        & \textbf{83.87} & 48.38 & 0.26 & 0.46 & 0.07 & 0.22 \\
Multi-Query & 48.39 & \textbf{54.83} & 0.18 & 0.35 & \textbf{0.11} & \textbf{0.25} \\ \hline
\end{tabular}
\caption{RAG results with different chunking methods and baseline settings: Llama 3.1 8B, KNN retrieval with top K=5 chunks, and LLM temperature of 0.5.}
\label{tab:query-transform-comparison}
\end{table}

Table \ref{tab:query-transform-comparison} illustrates the impact of different query transforms on the performance of the Llama 3.1 8B model across local and cross-industry tasks. The ``None" transform, representing no query modification, serves as a baseline, achieving 80.65\% accuracy for local MCQs and 52.94\% for cross-industry MCQs. It also shows relatively strong performance on local free text questions with BLEU and ROUGE scores of 0.41 and 0.60 respectively.\\

HyDE (Hypothetical Document Embeddings) demonstrates the best performance for local MCQs, improving accuracy to 83.87\%. However, it shows a slight decrease in cross-industry MCQ accuracy (48.38\%) compared to the baseline. HyDE's impact on free text metrics is mixed, with lower scores for local tasks but a marginal improvement in cross-industry BLEU.\\

The Multi-Query transform significantly underperforms in local MCQs (48.39\%), despite achieving the highest accuracy for cross-industry MCQs at 54.83\%. This suggests that the Multi-Query approach may be particularly effective for more complex queries requiring information synthesis across multiple domains (this is discussed further in Section \ref{query_transform_discussion}. Its free text performance is generally lower than the baseline, except for a slight improvement in cross-industry BLEU and ROUGE scores.

\subsection{Fine-tuning}

\begin{table}[H]
\centering
\begin{tabular}{lcc}
\hline
& \begin{tabular}[c]{@{}c@{}}MCQ\ Local\end{tabular} & \begin{tabular}[c]{@{}c@{}}MCQ\ Cross-Industry\end{tabular} \\ \hline
Llama 3.1 8B  & 0.46  & 0.48\\
Llama 3.1 40B & \textbf{0.58} & 0.50   \\
Llama 3.1 8B fine-tuned & 0.49  & \textbf{0.52}\\ \hline
\end{tabular}
\caption{Accuracy of the Llama 3.1 8B, Llama 3.1 40B, and fine-tuned Llama 3.1 8B models on MCQs, without RAG.}
\label{tab:fine-tuning}
\end{table}

Table \ref{tab:fine-tuning} presents the accuracy of three Llama 3.1 models on local and cross-industry multiple-choice questions (MCQs) without using RAG. The Llama 3.1 40B model demonstrates the highest accuracy (0.58) on local MCQs, while the fine-tuned Llama 3.1 8B model performs best on cross-industry MCQs with an accuracy of 0.52. Notably, the fine-tuned 8B model outperforms the base 8B model on both local and cross-industry questions, suggesting that domain-specific fine-tuning can enhance model performance. The 40B model shows superior performance on local questions, indicating that it is able to use its reasoning capabilities to deduce the correct answer. Nevertheless, this result is much lower than the performance of Llama 3.1 8B with RAG.

\subsection{Industry Classification}

\begin{table}[H]
\centering
\begin{tabular}{lcccc}
\hline
Model                    & Recall & Precision & Macro F1 score & Hamming Loss \\ \hline
RAG                      & 0.29   & 0.28      & 0.44     & 0.06         \\
Random Forest Classifier & \textbf{0.84}   & 0.45      & 0.59     & 0.02         \\
XGBoost                  & 0.62   & 0.77      & 0.69     & \textbf{0.01}         \\
MLP (Neural Network)     & 0.96   & 0.65      & 0.78     & \textbf{0.01}         \\
gpt-4o-mini              & 0.83   & \textbf{0.93}      & 0.86     & \textbf{0.01}         \\
gpt-4o                   & \textbf{0.84}   & \textbf{0.93}      & \textbf{0.87}     & \textbf{0.01}    \\ \hline     
\end{tabular}
\caption{Performance comparison of cross-industry classifiers.}
\label{tab:industry-classifiers}
\end{table}

Table \ref{tab:industry-classifiers} compares the performance of various industry classification models. The evaluation metrics include recall, precision, macro F1 score, and Hamming loss. The gpt-4o model achieves the highest overall performance with a macro F1 score of 0.87 and a Hamming loss of 0.01. The MLP (Neural Network) model demonstrates the highest recall (0.96), while gpt-4o and gpt-4o-mini show the highest precision (0.93). The RAG-based approach exhibits the lowest performance across all metrics, indicating that specialised classification models are more effective for this task. Among the machine learning models, XGBoost shows a balanced performance with high precision (0.77) and a low Hamming loss (0.01). These results suggest that both pre-trained language models and specialised machine learning approaches can be effective for industry classification, although LLM approaches are superior.

\subsection{Proposed Architecture}

\begin{table}[H]
\centering
\begin{tabular}{lcccccc}
\hline
\multirow{2}{*}{Pipeline} & \multicolumn{2}{c}{MCQ} & \multicolumn{4}{c}{Free text} \\ \cline{2-7} 
 & Local & Cross-Industry & \multicolumn{2}{c}{Local} & \multicolumn{2}{c}{Cross-Industry} \\  
 & Accuracy & Accuracy & BLEU & ROUGE & BLEU & ROUGE \\ \hline
Baseline        & 80.65 & 52.94 & 0.41 & 0.60 & 0.06 & 0.23 \\
Custom RAG & 85.32 & 72.15 & 0.45 & 0.61 & 0.13 & 0.26 \\
Fully LLM-based & \textbf{93.45} & \textbf{80.30} & \textbf{0.47} & \textbf{0.66} & \textbf{0.15} & \textbf{0.30} \\ \hline
\end{tabular}
\caption{Performance comparison of the custom RAG and LLM-based pipelines on MCQ and free-text questions. The baseline approach uses Llama 3.1 8B, and KNN retrieval with top K=5.}
\label{tab:proposed-pipelines}
\end{table}

The table \ref{tab:proposed-pipelines} provides a comparative analysis of the performance metrics for two different pipelines: the custom RAG pipeline and the fully LLM-based pipeline.\\ 

For MCQs, the custom RAG pipeline achieves an accuracy of 85.32\% for local questions and 72.15\% for cross-industry questions. The fully LLM-based pipeline demonstrates higher accuracy, with 93.45\% for local questions and 80.30\% for cross-industry questions. Most notably, both approaches achieve much higher performance on cross-industry questions than the baseline, and the LLM-based pipeline achieves an accuracy on cross-industry MCQs that is similar to the baseline performance on local questions.\\

On free-text questions, the custom RAG pipeline records BLEU and ROUGE scores of 0.45 and 0.61 for local queries, and 0.13 and 0.26 for cross-industry queries. The fully LLM-based pipeline achieves BLEU and ROUGE scores of 0.47 and 0.66 for local queries, and 0.15 and 0.30 for cross-industry queries. BLEU and ROUGE scores of the custom pipelines show a lower improvement than accuracy on MCQs, though this is a result of the metrics themselves, as discussed in Section \ref{models_discussion}. \\

Overall, the fully LLM-based pipeline achieves the best performance across all types of questions.

\section{Discussion}

\subsection{RAG}
\subsubsection{Models}\label{models_discussion}

Model size and capability have a strong impact on RAG performance. The smallest model, Google Gemma 2B, is most often unable to answer the question posed, and when it does provide an answer it is usually wrong. Llama 2 13B is an older, less capable model, and although it is bigger than the newer 8B models, it struggles to answer questions directly and recites the source content instead, showing its weak reasoning capabilities. These issues are particularly pronounced for more complicated cross-industry questions.\\

Listing \ref{lst:models_qa_example} below shows a local multi-hop question example with the RAG responses given by each model. The correct answer must contain two parts: the first part requires extracting information from a Table, while the second is verbatim text from the source document. Most models (except Gemma 2B) got the second part correct. However, only the fine-tuned model gets the first part of the answer correct, showcasing that certain knowledge has been embedded into the fine-tuned model and is being used successfully when combined with RAG methods.\\

Also notable is that all models contain slightly different phrasings of the same answer, containing additional words/details to various extents. Even though multiple phrasings can be correct, BLEU and ROUGE metrics are sensitive to exact phrasing, resulting in low BLEU and ROUGE scores, particularly for more complicated cross-industry questions. Local questions, on the other hand, can more successfully be evaluated using BLEU and ROUGE as they are often shorter and more factual, so a correct answer will score above 0.9 on both metrics.

\begin{lstlisting}[language=json,firstnumber=1,label={lst:models_qa_example},caption={Example local multi-hop question with RAG answers using different models}]
Question: "What is the category of the 'Payload fuel economy' metric in the Health Care Distributors industry and how should this metric be calculated?"

Correct answer: "The category of the 'Payload fuel economy' metric is Quantitative. Payload fuel economy shall be calculated as: total gallons of fuel consumed / revenue tons miles (RTM), where revenue ton miles (RTM) = total weight of paid tonnage transported (payload) x total distance in miles goods were transported."

Google Gemma 2B answer: "The category of the 'Payload fuel economy' metric in the Health Care Distributors industry is **Transportation**.\n\nThe metric should be calculated as:\n\n```\nPayload fuel economy = total gallons of fuel consumed / revenue tons miles (RTM)\n```\n\nwhere:\n\n* **total gallons of fuel consumed** is the total amount of fuel consumed by the entity's transportation fleet.\n* **revenue tons miles (RTM)** is the total weight of paid tonnage transported (payload) multiplied by the total distance in miles goods were transported."
"bleu_score": 0.36
"rouge_l_score": 0.65

Llama 2 13B answer: "Based on the context information provided, the 'Payload fuel economy' metric in the Health Care Distributors industry falls under the category of 'Fleet Fuel Management' and should be calculated as the total gallons of fuel consumed divided by the revenue tons miles (RTM) of the delivery fleet, where RTM is the weight of paid tonnage transported (payload) multiplied by the total distance in miles goods were transported. This calculation should be aggregated for types of transportation, including air, marine, rail, and road."
"bleu_score": 0.30
"rouge_l_score": 0.50

Llama 3.1 8B answer: "The category of the 'Payload fuel economy' metric is HC-DI-110a.1. This metric should be calculated as total gallons of fuel consumed divided by revenue ton miles (RTM), where revenue ton miles (RTM) is the total weight of paid tonnage transported multiplied by the total distance in miles goods were transported."
"bleu_score": 0.60
"rouge_l_score": 0.75

Llama 3.1 8B fine-tuned answer: "The category of the 'Payload fuel economy' metric is Quantitative. This metric should be calculated as total gallons of fuel consumed divided by revenue ton miles (RTM), where revenue ton miles (RTM) is the total weight of paid tonnage transported multiplied by the total distance in miles goods were transported."
"bleu_score": 0.65
"rouge_l_score": 0.77

Llama 3 70B answer: "The category of the 'Payload fuel economy' metric in the Health Care Distributors industry is HC-DI-110a.1. This metric should be calculated as total gallons of fuel consumed divided by revenue tons miles (RTM), where revenue ton miles (RTM) equals total weight of paid tonnage transported (payload) multiplied by total distance in miles goods were transported."
"bleu_score": 0.61
"rouge_l_score": 0.74

Mixtral 8x7B answer: "The 'Payload fuel economy' metric in the Health Care Distributors industry falls under the category of Fleet Fuel Management. This metric should be calculated as the total gallons of fuel consumed divided by revenue tons miles (RTM), where RTM is equal to the total weight of paid tonnage transported (payload) multiplied by the total distance in miles goods were transported. Payload includes the weight of paid tonnage and excludes the vehicle weight. The entity should aggregate payload fuel economy for different types of transportation, including air, marine, rail, and road transportation."
"bleu_score": 0.31
"rouge_l_score": 0.56
\end{lstlisting}

\subsubsection{Data Chunking} \label{chunking_discussion}

A notable observation from Table \ref{tab:chunking} is that combining large fixed-sized chunks (Sentence 1024) with top-5 retrieval hurts performance on MCQs, but not on free-text questions. This suggests that the retrieved context confuses the LLM when it has to choose between five similar answer options, but not when the LLM writes its own response to the question in free text format. Interestingly, this is not a problem for semantic/markdown chunking, which also contains large information chunks. This is because, even though the chunks are large, they are semantically coherent, whereas fixed-size chunks do not contain complete information as text is simply cut off at fixed points. \\

Semantic and custom markdown chunking methods yield the best results, as the context given to the LLM is coherent. The custom markdown adds the additional benefit of isolating tables, so more specific context can be retrieved for questions that ask for specific information from tables.

\subsubsection{Retrieval Techniques}

While both KNN and hybrid retrievers yield similar performance, there is a notable drop in accuracy on cross-industry MCQs using the hybrid retriever. This could suggest that incorporating text-based keyword search does not provide a benefit for the sustainability reporting domain, and may indeed detract from the power of embedding similarity.

\subsubsection{Query Transforms}\label{query_transform_discussion}

Hyde exhibits 4\% better performance on local MCQ accuracy than the baseline, but 8.6\% lower performance on cross-industry performance. This suggests that the hypothetical generated document is often accurate for factual questions, but struggles for more complex queries based on multiple source documents. This is evident in Listing \ref{lst:query_transforms_example}, where HyDE's answer to a cross-industry question contains hallucinated content such as the ``FTSE Nareit Classification Structure". \\

Multi-query exhibits a 40\% reduction in accuracy on local answers and a 3.6\% improvement in cross-industry accuracy relative to baseline. Its superior ability to handle complex questions is shown in its highly accurate answer in Listing \ref{lst:query_transforms_example}. However, the process of rephrasing questions in multiple ways harms the response to specific factual questions that are sensitive to detail and phrasing. This is shown in Listing \ref{lst:query_transforms_local}, where the baseline query gets the correct answer, but rephrasing the query multiple times induces hallucination in the last part of the answer. The fall in quality of simple factual responses is too great to justify the advantages of multi-query transforms for answering complex questions.
\newpage
\begin{lstlisting}[language=json,firstnumber=1,label={lst:query_transforms_local},caption={Example local single-hop question with RAG answers using different query transforms}]
Question: "What are some examples of efforts to reduce the environmental impact of logistics for the Health Care Distributors industry?"

Correct answer: "Relevant efforts to describe include, but are not limited to, upgrades to fleet (fuel efficiency), usage of alternative and/or renewable fuels, optimized logistics routes, and idling reduction programs.",

Baseline answer: "Efforts to reduce the environmental impact of logistics for the Health Care Distributors industry may include upgrading to more fuel-efficient fleets, using alternative and/or renewable fuels, optimizing logistics routes, and implementing idling reduction programs."
"bleu_score": 0.34
"rouge_l_score": 0.57

Multi-query answer: "Efforts to reduce the environmental impact of logistics for the Health Care Distributors industry may include upgrading to more fuel-efficient fleets, using alternative and/or renewable fuels, optimizing logistics routes, implementing idling reduction programs, partnering with energy-efficient suppliers, and investing in smart logistics technologies to enhance route efficiency and decrease fuel consumption.",
"bleu_score": 0.13
"rouge_l_score": 0.28

HyDE answer: "Health care distributors can implement various strategies to reduce the environmental impact of their logistics operations. For instance, they can optimize their transportation routes and modes to minimize fuel consumption and emissions. This might involve using more fuel-efficient vehicles, such as electric or hybrid trucks, or switching to rail or sea transportation for longer distances. Additionally, distributors can work with suppliers to reduce packaging materials and implement sustainable packaging practices, such as using biodegradable or recyclable materials. They can also explore alternative fuel options, like biofuels or hydrogen fuel cells, to power their fleets. Furthermore, distributors can invest in route optimization software to reduce idling time and lower emissions. By implementing these strategies, health care distributors can not only reduce their environmental footprint but also improve their bottom line by reducing fuel costs and increasing efficiency.",
"bleu_score": 0.004
"rouge_l_score": 0.115
\end{lstlisting}

\begin{lstlisting}[language=json,firstnumber=1,label={lst:query_transforms_example},caption={Example cross-industry single-hop question with RAG answers using different query transforms}]
"question": "Which industry has more 'Quantitative' metrics to disclose in Table 1 between Real Estate Services and Agricultural Products industries?"

"correct_answer": "The Agricultural Products industry has more 'Quantitative' metrics to disclose, with seven metrics in Table 1 compared to the Real Estate Services industry, with three metrics."

No transform answer: "Based on the provided context, it appears that the Agricultural Products industry has more 'Quantitative' metrics to disclose in Table 1. This is evident from the detailed metrics listed under the \"Metrics\" section, which include specific calculations and thresholds for disclosure, such as FB-AG-440a.1 and FB-AG-110a.1. In contrast, there is no mention of specific metrics or disclosure requirements for the Real Estate Services industry in the provided context."

Multi-query answer: "The Agricultural Products industry has more 'Quantitative' metrics to disclose in Table 1 compared to the Real Estate Services industry."

HyDE answer: "The Real Estate industry has 4 'Quantitative' metrics to disclose in Table 1, specifically Leasable floor area, by property subsector, Percentage of indirectly managed assets, by property subsector, and Average occupancy rate, by property subsector. These metrics are aligned with the FTSE Nareit Classification Structure and are disclosed separately for each portion of the entity's portfolio where properties are classified into subsectors."
\end{lstlisting}

\subsubsection{Industry Retrieval}

A consistent theme observed in the RAG results across all methods is that performance on cross-industry questions is lower than on local ones. This phenomenon was manually investigated by examining the chunks retrieved for cross-industry queries. It was found that the retriever gets chunks from multiple industries that could be related to the query, thus confusing the LLM that generates the answer. One such example question is shown in Listing \ref{lst:industry_query_example} below along with the retrieved industries using the baseline RAG method, as well as those classified by the LLM industry classifier. The baseline RAG selects five chunks from four different related industries, where the target is two industries. The LLM classifier identifies the correct industries as being the highest probability ones intended by the user.

\begin{lstlisting}[language=json,firstnumber=1,label={lst:industry_query_example},caption={Example cross-industry single-hop question}]
"question": "What are the activity metrics for the car transport and air transport industries?"
"target industries": "b61-airlines", "b63-automobiles"
"industries retrieved by baseline RAG": "b62-auto-parts" (x2), "b63-automobiles", "b60-air-freight-and-logistics", "b61-airlines"
"industries output by the LLM classifier": "b63-automobiles", "b61-airlines"
\end{lstlisting}

\subsection{Proposed Architecture}

Both custom pipelines demonstrate significant improvements in accuracy on cross-industry questions, and this can be largely attributed to the integration of the LLM industry classifier, which constrains the domain of knowledge from which information is retrieved for a query, thus increasing the likelihood of retrieving the correct chunks. \\

Only the LLM-based pipeline, however, reaches performance on cross-industry questions that is comparable to performance on local ones, despite both architectures using the same industry classifier. This is explained by the difference in chunking and retrieval methods adopted in the two approaches. The RAG pipeline still utilises information that has been `generically' segmented, rather than chunks that are customised to the query itself. Furthermore, top-5 retrieval is used rather than allowing an LLM to decide on the appropriate number of chunks based on their context. While the top-5 retrieval hurdle can be tackled using more sophisticated techniques, such as by setting a similarity threshold and selecting all chunks above this, selecting this threshold is complex, requiring thorough experimentation, and there is unlikely to be a single threshold that is appropriate for all query types. Finally, the RAG-based pipeline relies on embedding similarity to select chunks, which was the best-performing retriever tested.\\

The LLM-based pipeline retrieves the most customised context for answering the specific query, lessening the burden on the generator LLM of utilising its own reasoning capabilities to understand the information and generate an answer. It also minimises the chances of irrelevant context being retrieved (which is a problem when a specific number of chunks must be retrieved at each query). Nevertheless, this architecture is not without weaknesses. Most notably, the increased number of LLM calls may increase the chance of hallucination of the retrieved content. Furthermore, there seems to be a limit to its performance on complex questions still, as the accuracy on cross-industry questions is 80.30\% rather than being closer to the 93.25\% accuracy on local MCQs. This suggests there exist limits to the reasoning capabilities of the generator LLM, despite being provided with improved context.

\chapter{Conclusion and Future Work}

\section{Conclusion}

This research was driven by a need to support companies in navigating IFRS reporting standards for the preparation of compliant sustainability reports. Two major gaps were identified in this domain that impeded the creation of an AI assistant for this task: (1) the lack of a high-quality question-answer dataset, and (2) the lack of a RAG system customised for question-answering in this domain. This work addresses both of these gaps. \\

This project utilises the official reports published by IFRS to construct, from scratch, a dataset and two question-answering systems in the sustainability reporting domain. A novel LLM-based pipeline is developed to generate and evaluate domain-specific synthetic question-answer pairs. This pipeline is used to produce a dataset of 1,063 diverse, well-referenced, domain-specific questions and answers. \\

The dataset was used to evaluate RAG methods for a domain-specific chatbot and design two custom architectures that improve on standalone RAG methods. Specifically, the dataset was used to fine-tune a model for question-answering and evaluate all the methods tested on various levels of complexity. The dataset was also used to train a range of industry classifiers, though ultimately a prompt-based classifier displayed the best performance. The prompt-based industry classifier and fine-tuned generation model were incorporated into two pipelines - a RAG-based pipeline and a fully LLM-based architecture. The fully LLM-based architecture performed best, although both implementations showed significant improvements in performance over the baseline. \\

Nevertheless, the methods presented in the project present some limitations:

\begin{itemize}
    \item \textbf{QA generation prompts:} All prompts in the project are designed using generalisable techniques such few-shot and chain of thought prompting, and templates are used throughout to act as LLM ``functions" to standardise the overall pipeline. Nevertheless, to obtain high quality questions for the sustainability reporting domain, the details in the prompts had to be carefully crafted and are heavily related to the underlying context of the chatbot. To adapt the QA generation pipeline to other specialised domains, the prompts will likely need to be edited to accommodate domain-specific concerns.
    \item \textbf{QA evaluation:} The QA evaluation methods are constructed using LLMs, which may fail to evaluate questions at the human level or may themselves hallucinate. Question quality was checked manually throughout the project as methods were iterated, though this was not done on all 1,063 questions presented in the final dataset.
    \item \textbf{QA post-processing:} Due to cost considerations, the custom post-processing functions - quality improvement and question generalisation - were implemented qualitatively on a small sample rather than applied to the entire dataset. Nevertheless, the generated dataset displays high quality even before post-processing.
    \item \textbf{Chatbot evaluation:} The knowledge of the RAG system was tested and the chatbot was constrained to its specific domain. However, the chatbot itself has not been evaluated through the lens of user experience and long-context dialogue.
\end{itemize}

\section{Future Work}

Future work can build on the contributions of this project in several ways to address the limitations identified as well as to expand on the methods proposed. The following work is proposed.
\begin{itemize}
    \item \textbf{Adapting methods to different domains:} The designed architecture can be adapted to different domains by editing the prompts where domain-specific context is most relevant.
    \item \textbf{Human QA evaluation:} Human evaluation on the question dataset can be conducted to verify the quality of the generated questions and answers. This can be done by domain experts and/or people with an understanding of IFRS sustainability reporting standards. Evaluation must be done systematically, with each question being graded on a set of quality metrics.
    \item \textbf{Mass post-processing:} The investigation of the custom post-processing functions showed that they are effective, and would merit an application across the entire dataset on questions that do score below the threshold on quality metrics.
    \item \textbf{Human chatbot evaluation:} Human evaluation of the chatbot can be conducted. This would involve defining the key metrics on which the chatbot must perform well. Some examples include usefulness, context awareness, factual accuracy, and politeness, measured on Likert or other numerical scales to increase comparability across human evaluators. An alternative to human evaluation of the chatbot is to prompt other LLMs to act as typical users and instruct them to rate their experience from this perspective.
    \item \textbf{Cross-industry multi-query:} A custom RAG pipeline can be designed to leverage multi-query transforms only on cross-industry questions, where it displays advantages over the baseline.
    \item \textbf{Leveraging the QA dataset:} The dataset can be used to train and evaluate further advanced RAG systems and other models within the sustainability reporting domain. 
    \item \textbf{Generating more questions:} The QA generation and evaluation pipeline can be employed to generate questions within this domain on a larger scale.
\end{itemize}

\begin{appendices}

\chapter{Additional Results} \label{appxa}

\section{Open-source Evaluation Framework Examples}

Below is an example question whose answer is untrue based on the source document, which does not provide a percentage. The question is also irrelevant from the perspective of a company following the reporting guidelines as it directly asks about some entity rather than how it should report on this metric.

\begin{lstlisting}[language=json,firstnumber=1,label={lst:opensource_eval},caption={Sample MCQ generated using the baseline prompt method}]
"question": "What is the entity's average refrigerant emissions rate?",
    "optionA": "5%",
    "optionB": "10%",
    "optionC": "15%",
    "optionD": "20%",
    "optionE": "25%",
    "answer": "C",
    "industries": [
      "b22-food-retailers-and-distributors"
    ]
\end{lstlisting}

Using LlamaIndex (which uses binary scores), this question was given a score of 1 in both faithfulness and relevance (i.e. the highest score). DeepEval (which uses continuous scores from 0 to 1) scored it 0.9 on faithfulness and 1.0 on relevance. Our custom framework, which contains extra prompt context for how it should measure faithfulness and relevance, scored it 1 in faithfulness and 4 in relevance.

\section{Retrieval Techniques}\label{retriever_appx}
\begin{table}[H]
\centering
\begin{tabular}{lcccccc}
\hline
\multirow{3}{*}{Retriever} & \multicolumn{2}{c}{MCQ} & \multicolumn{4}{c}{Free text} \\ \cline{2-7} 
 & Local & Cross-Industry & \multicolumn{2}{c}{Local} & \multicolumn{2}{c}{Cross-Industry} \\  
 & Accuracy & Accuracy & BLEU & ROUGE & BLEU & ROUGE \\ \hline
Semantic hybrid   & 45.16 & 27.43 & 0.07 & 0.24 & 0.05 & 0.17 \\
SVM               & 41.94 & 37.28 & 0.06 & 0.24 & 0.06 & 0.18 \\
Linear regression & 45.16 & 37.25 & 0.07 & 0.24 & 0.06 & 0.20 \\
MMR               & 35.48 & 27.45 & 0.06 & 0.24 & 0.06 & 0.19 \\ \hline
\end{tabular}
\caption{RAG results with different chunking methods and baseline settings: Llama 3.1 8B, retrieval of top K=5 chunks, no query transform, and LLM temperature of 0.5.}
\label{tab:retriever-comparison-appx}
\end{table}

\chapter{Prompts and Schemas}\label{appxb}

\section{PDF Parsing Prompts}

\begin{lstlisting}[language=json,firstnumber=1,label={lst:pdf_parse_prompt},caption={Schemas and prompts for PDF table and text extraction}]
# Tool for processing table pages
table_pages_processor = {
    "name": "process_table_pages",
    "description": "Process the pages containing tables and extract specified information",
    "input_schema": {
        "type": "object",
        "properties": {
            "sustainability_metrics_table": {
                "type": "string",
                "description": "Markdown of sustainability metrics table",
            },
            "activity_metrics_table": {
                "type": "string",
                "description": "Markdown of activity metrics table, if present",
            },
            "report_title": {"type": "string", "description": "Title of the report"},
            "industry": {"type": "string", "description": "Industry of the report"},
        },
        "required": ["sustainability_metrics_table", "report_title", "industry"],
    },
}

# Prompt for processing tables
table_prompt = "Process these PDF pages and extract the following information:\n"
    "1. Markdown of the entire sustainability metrics table, which has these five columns: TOPIC, METRIC, CATEGORY, UNIT OF MEASURE, CODE (make sure to exclude any text that has been crossed out, but include any underlined text!)\n"
    "2. Markdown of the activity metrics table, if present, which has these four columns: ACTIVITY METRIC, CATEGORY, UNIT OF MEASURE, CODE (make sure to exclude any text that has been crossed out, but include any underlined text!). If there is no activity metrics table, omit this field.\n"
    "3. Report title\n"
    "4. Industry\n\n"
    "Exclude any headers (APPENDIX B OF DRAFT IFRS S2 CLIMATE-RELATED DISCLOSURES) or (EXPOSURE DRAFT MARCH 2022) and footers (2022 SASB, part of Value Reporting Foundation. All rights reserved.) of the pages, but include footnotes. Exclude any instances of 'continued...' or '...continued'. Exclude any text that has been crossed out, but include any underlined text. Note that tables may continue across pages. Provide the output in the specified schema."

# Tool for processing individual pages
individual_page_processor = {
    "name": "process_individual_page",
    "description": "Process a single page of a PDF document and extract specified information",
    "input_schema": {
        "type": "object",
        "properties": {
            "text_content": {
                "type": "string",
                "description": "All text content, specifically in markdown format, excluding the tables Table 1. Sustainability Disclosure Topics & Metrics and Table 2. Activity Metrics." ,
            },
            "page_number": {
                "type": "integer",
                "description": "Page number of the processed page in the footer",
            },
        },
        "required": ["text_content", "page_number"],
    },
}

# Prompt for processing text
prompt = f"""Process this PDF page and extract all text content, excluding the tables Table 1. Sustainability Disclosure Topics & Metrics and Table 2. Activity Metrics, but including any other tables that appear. Exclude any text that has been crossed out, but include any underlined text. Exclude any headers (APPENDIX B OF DRAFT IFRS S2 CLIMATE-RELATED DISCLOSURES) or (EXPOSURE DRAFT MARCH 2022) and footers (2022 SASB, part of Value Reporting Foundation. All rights reserved.) of the pages, but include footnotes. Exclude the section header titled 'Sustainability Disclosure Topics & Metrics'. Exclude subsections 'Table 1. Sustainability Disclosure Topics & Metrics' and 'Table 2. Activity Metrics'. Exclude any instances of 'continued...' or '...continued'. The report title is '{report_title}', the industry is '{industry}' (but DO NOT include the report title and industry as sections in the markdown). Provide the output in the specified schema."""

\end{lstlisting}

\begin{lstlisting}[language=json,firstnumber=1,label={lst:mcq_schema},caption={Multiple-choice question output schema}]
{    
    "name": "qa_pair_schema",
    "description": "Generate multiple choice question-answer pairs from industry markdown",
    "input_schema": {
        "type": "object",
        "properties": {
            "qa_pairs": {
                "type": "array",
                "items": {
                    "type": "object",
                    "properties": {
                        "question": {"type": "string", "description": "the question"},
                        "optionA": {"type": "string", "description": "option A"},
                        "optionB": {"type": "string", "description": "option B"},
                        "optionC": {"type": "string", "description": "option C"},
                        "optionD": {"type": "string", "description": "option D"},
                        "optionE": {"type": "string", "description": "option E"},
                        "answer": {"type": "string", "description": "the correct answer option letter"},
                        "reference_text": {
                            "type": "array",
                            "items": {"type": "string"},
                            "description": "the verbatim text taken directly from the report that is used to generate the question and correct answer"
                        },
                        "pages": {
                            "type": "array",
                            "items": {"type": "string"},
                            "description": "List of page numbers"
                        },
                    },
                    "required": [ "question", "optionA", "optionB", "optionC", "optionD", "optionE", "answer", "reference_text", "pages"]
                }
            }
        },
        "required": ["qa_pairs"]
    }
}
\end{lstlisting}

\section{Industry Pair Generation Prompt}

\begin{lstlisting}[language=json,firstnumber=1,label={lst:memprompt},caption={Prompt for the generation of industry pairs for cross-industry questions, based on existing industry groups and industry descriptions taken from the markdown}]
prompt = f"""As a specialist consultant on IFRS sustainability reporting standards, please suggest 5 groups of 5 different industries that are most likely to come up when considering reporting standards. These should be industries where comparisons or relationships in sustainability reporting would be particularly relevant or insightful.

Use the following industry groups and descriptions as a reference:

[insert industry groups and industry descriptions here]

For each suggestion, provide:
1. The industries involved (using their codes, e.g., b1-apparel-accessories-and-footwear, b2-appliance-manufacturing, etc.)
2. A brief explanation of why these industries are relevant to compare in terms of sustainability reporting.

Use the provided schema to format your response."""
\end{lstlisting}

\section{Question Structures for Few-shot Prompting QA Generation}

\begin{lstlisting}[language=json,firstnumber=1,label={lst:qa_structures},caption={User-representative sample questions for `CoT + Few-shot' question generation}]
{
    "Local": {
      "single_hop": [
        "On what page can I find details about xxx for the xxx industry?",
        "Can you provide for me the unit of measure I should use for the xxx metric in the xxx industry?",
        "Can you provide for me the unit of measure I should use for the xxx metric in the xxx industry (as listed in the relevant table)?",
        "What is the category of xxx metric in the xxx industry?",
        "What is the category of xxx metric in the xxx industry (as listed in the relevant table)?",
        "What is the code of xxx metric in the xxx industry?",
        "What is the code of xxx metric in the xxx industry (as listed in the relevant table)?",
        "Can you tell me what topics are covered in the xxx section of the xxx document?",
        "For the listed xxx requirement, what does a company in the xxx industry need to do?",
        "For the listed xxx requirement, what metric should a company in the xxx industry use for xxx?"
      ],
      "multi_hop": [
        "Can you provide for me the unit of measure I should use for each of the metrics for 'xxx' topic(s) in the xxx industry?",
        "What is the category of xxx metric in the xxx industry and how should this metric be calculated?",
        "What is the code of xxx metric in the xxx industry and how should this metric be calculated?",
        "Can you tell me if I reported this topic and/or metric correctly for the xxx industry?",
        "Can you provide he unit of measure (as listed in the relevant table) for all the metrics for xxx topic(s) in the xxx industry?",
        "Can you provide he unit of measure (as listed in the relevant table) for all the metrics for xxx topic(s) in the xxx industry and how should these metrics be calculated?",
        "What is the category of xxx metric in the xxx industry and how should this metric be disclosed?",
        "What is the code of xxx metric in the xxx industry and how should this metric be disclosed??",
        "Can you tell me what topics are covered in the xxx section of the xxx document and what are the key takeaways I need to know as a company writiny my sustainability report in the xxx industry?",
        "For the listed xxx requirement, what metric should a company in the xxx industry use for xxx and how should this metric be calculated?"
      ]
    },
    "Cross-industry": {
      "single_hop": [
        "What topics should I report on for xxx and xxx (and xxx...) industries?",
        "Give me all the metrics from Table 1 that are in category 'Discussion and Analysis' for xxx and xxx (and xxx...) industries.",
        "Give me all the metrics from Table 1 that are in category 'Quantitative' for xxx and xxx (and xxx...) industries.",
        "Give me all the metrics that are in category 'Discussion and Analysis' for xxx and xxx (and xxx...) industries.",
        "Give me all the metrics that are in category 'Quantitative' for xxx and xxx (and xxx...) industries.",
        "What are the common metrics shared between xxx and xxx (and xxx...) industries, if any?",
        "List the activity metrics for xxx and xxx (and xxx...) industries.",
        "Compare the number of quantitative metrics between xxx and xxx (and xxx...) industries.",
        "Which industry has more 'Quantitative' metrics to disclose in Table 1 between xxx and xxx (and xxx) industries?",
        "Which industry has more 'Qualitative' metrics to disclose in Table 1 between xxx and xxx (and xxx) industries?"
      ],
      "multi_hop": [
        "Give me the codes for all the 'Quantitative' metrics in Table 1 for xxx and xxx (and xxx...) industries.",
        "Give me the codes for all the 'Discussion and Analysis' metrics in Table 1 for xxx and xxx (and xxx...) industries.",
        "Give me the codes for all the 'Quantitative' metrics for xxx and xxx (and xxx...) industries.",
        "Give me the codes for all the 'Discussion and Analysis' metrics for xxx and xxx (and xxx...) industries.",
        "What units of measurement are used across xxx and xxx (and xxx...) industries for measuring xxx?",
        "Are there any common sustainability topics between xxx and xxx (and xxx...) industries? If so, how do their reporting requirements differ?",
        "When preparing sustainability reports, how do the metrics for measuring xxx differ between the xxx and xxx (and xxx...) industries?",
        "How does the calculation of xxx-related metrics differ between xxx and xxx (and xxx...) companies?",
        "When calculating xxx metrics, what is the key difference between xxx and and xxx (and xxx...) companies?",
        "Analyze the reporting complexity of xxx and xxx (and xxx...) industries by comparing the number of required metrics and the diversity of measurement units used",
        "Comparing the xxx and xxx (and xxx...) industries, which sustainability topic is more prominently addressed in the xxx industry but not explicitly covered for xxx industry?"
      ]
    },
    "Free_local":{
      "single_hop":[
        "What does the IFRS require companies to disclose regarding xxx for the xxx industry?",
        "Describe the calculation method for the xxx metric for the xxx industry.",
        "How should I report xxx for the xxx industry?",
        "How should I calculate xxx for the xxx industry?",
        "What are some examples of xxx? for the xxx industry?",
        "What targets do I need to discuss for xxx for the xxx industry?",
        "Are there any specific programs or regulations I need to discuss for the xxx industry? Which ones?",
        "What are the factor(s) that the company should describe in its approach and strategy to xxx, according to the xxx metrics for the xxx industry?",
        "On what page can I find details about xxx for the xxx industry?",
        "Can you provide for me the unit of measure I should use for the xxx metric in the xxx industry?",
        "Can you provide for me the unit of measure I should use for the xxx metric in the xxx industry (as listed in the relevant table)?",
        "What is the category of xxx metric in the xxx industry?",
        "What is the category of xxx metric in the xxx industry (as listed in the relevant table)?",
        "What is the code of xxx metric in the xxx industry?",
        "What is the code of xxx metric in the xxx industry (as listed in the relevant table)?",
        "Can you tell me what topics are covered in the xxx section of the xxx document?",
        "For the listed xxx requirement, what does a company in the xxx industry need to do?",
        "For the listed xxx requirement, what metric should a company in the xxx industry use for xxx?"
      ],
      "multi_hop":[
        "Compare and contrast the reporting requirements for xxx and xxx topics in the xxx industry.",
        "Analyze the reporting complexity of xxx and xxx (and xxx...) industries by comparing the number of required metrics.",  
        "If I have xxx, how should I report it?",
        "What do I need to discuss or disclose for xxx topic?",
        "Can you provide for me the unit of measure I should use for each of the metrics for 'xxx' topic(s) in the xxx industry?",
        "What is the category of xxx metric in the xxx industry and how should this metric be calculated?",
        "What is the code of xxx metric in the xxx industry and how should this metric be calculated?",
        "Can you tell me if I reported this topic and/or metric correctly for the xxx industry?",
        "Can you provide he unit of measure (as listed in the relevant table) for all the metrics for xxx topic(s) in the xxx industry?",
        "Can you provide he unit of measure (as listed in the relevant table) for all the metrics for xxx topic(s) in the xxx industry and how should these metrics be calculated?",
        "What is the category of xxx metric in the xxx industry and how should this metric be disclosed?",
        "What is the code of xxx metric in the xxx industry and how should this metric be disclosed??",
        "Can you tell me what topics are covered in the xxx section of the xxx document and what are the key takeaways I need to know as a company writiny my sustainability report in the xxx industry?",
        "For the listed xxx requirement, what metric should a company in the xxx industry use for xxx and how should this metric be calculated?"
      ]
    },
    "Free_cross-industry":{
      "single_hop": [
        "What topics should I report on for xxx and xxx (and xxx...) industries?",
        "Give me all the metrics from Table 1 that are in category 'Discussion and Analysis' for xxx and xxx (and xxx...) industries.",
        "Give me all the metrics from Table 1 that are in category 'Quantitative' for xxx and xxx (and xxx...) industries.",
        "Give me all the metrics that are in category 'Discussion and Analysis' for xxx and xxx (and xxx...) industries.",
        "Give me all the metrics that are in category 'Quantitative' for xxx and xxx (and xxx...) industries.",
        "What are the common metrics shared between xxx and xxx (and xxx...) industries, if any?",
        "List the activity metrics for xxx and xxx (and xxx...) industries.",
        "Compare the number of quantitative metrics between xxx and xxx (and xxx...) industries.",
        "Which industry has more 'Quantitative' metrics to disclose in Table 1 between xxx and xxx (and xxx) industries?",
        "Which industry has more 'Qualitative' metrics to disclose in Table 1 between xxx and xxx (and xxx) industries?"
      ],
      "multi_hop": [
        "Give me the codes for all the 'Quantitative' metrics in Table 1 for xxx and xxx (and xxx...) industries.",
        "Give me the codes for all the 'Discussion and Analysis' metrics in Table 1 for xxx and xxx (and xxx...) industries.",
        "Give me the codes for all the 'Quantitative' metrics for xxx and xxx (and xxx...) industries.",
        "Give me the codes for all the 'Discussion and Analysis' metrics for xxx and xxx (and xxx...) industries.",
        "What units of measurement are used across xxx and xxx (and xxx...) industries for measuring xxx?",
        "Are there any common sustainability topics between xxx and xxx (and xxx...) industries? If so, how do their reporting requirements differ?",
        "When preparing sustainability reports, how do the metrics for measuring xxx differ between the xxx and xxx (and xxx...) industries?",
        "How does the calculation of xxx-related metrics differ between xxx and xxx (and xxx...) companies?",
        "When calculating xxx metrics, what is the key difference between xxx and and xxx (and xxx...) companies?",
        "Analyze the reporting complexity of xxx and xxx (and xxx...) industries by comparing the number of required metrics and the diversity of measurement units used",
        "Comparing the xxx and xxx (and xxx...) industries, which sustainability topic is more prominently addressed in the xxx industry but not explicitly covered for xxx industry?"
      ]
    },
}
\end{lstlisting}
\newpage
\section{Post-processing functions}

\begin{lstlisting}[language=json,firstnumber=1,label={lst:quality_improvement_prompt},caption={Quality improvement function prompt}]
"system" = "You are an expert at improving questions while maintaining their core meaning and structure."

"user": f"""
Improve the following question by focusing on the {metric_to_improve} metric. 
The question is from the {', '.join(question['industries'])} industry/industries.

Original question: {question['question']}

Current metrics:
Faithfulness: {question['faithfulness_score']}
Relevancy: {question.get(f"relevancy_score_{question['industries'][0]}", 'N/A')}
Specificity: {question['specificity_score']}

Relevant industry content:
{relevant_content}

Guidelines for improvement:
1. If improving faithfulness: Ensure accuracy, verify metrics and concepts, remove misleading information.
2. If improving relevancy: Focus on industry-specific aspects, use appropriate terminology.
3. If improving specificity: Add precise details, use exact metric names or values, narrow the scope if needed.

Maintain the original structure, intent, and difficulty level of the question.
If it's a multi-choice question, preserve that format.

Provide only the improved question.
"""
\end{lstlisting}
\newpage
\begin{lstlisting}[language=json,firstnumber=1,label={lst:generalisation_prompt},caption={Generalisation function prompt}]
"system" = "You are an expert at rephrasing questions to make them slightly more vague while maintaining their core meaning."

"user" = f"""
Refine the following question by ONLY slightly generalizing the name of the industry. 
The question is from the {industry} industry. 
Keep the entire question exactly the same, only modifying the industry name to be slightly more general.

Original question: {question}

Guidelines:
1. ONLY change the specific industry name to a slightly more general term.
2. Keep ALL other parts of the question, including technical terms, metrics, and structure, exactly the same.
3. The refined question should be identical to the original except for the industry name.
4. Make the change in industry name as minimal as possible while still generalizing slightly.

Examples:
1. Original: What is the code for the 'Gross global Scope 1 emissions, percentage covered under emissions-limiting regulations' metric in the Coal Operations industry?
   Refined: What is the code for the 'Gross global Scope 1 emissions, percentage covered under emissions-limiting regulations' metric in the industry about operating coal?

2. Original: In the Apparel, Accessories & Footwear industry, what percentage of raw materials should be third-party certified to environmental or social sustainability standards?
   Refined: For clothing companies, what percentage of raw materials should be third-party certified to environmental or social sustainability standards?

3. Original: What is the reporting metric for water consumption in Oil & Gas Exploration & Production operations?
   Refined: What is the reporting metric for water consumption in fossil fuel extraction operations?

4. Original: For Electric Utilities, what is the RIF (Recordable Incident Frequency) safety performance indicator?
   Refined: For power utility companies, what is the RIF (Recordable Incident Frequency) safety performance indicator?

Provide only the revised question without any explanation.
"""

\end{lstlisting}

\newpage
\begin{lstlisting}[language=json,firstnumber=1,label={lst:sba_prompt},caption={SBA check prompt}]
system = "You are an expert question designer. Given a question and full context, you check if it has one and only one correct option."

prompt = f"""
Full context:
{context}
Based on the following question and the given context, determine if there is only one correct answer option, by identifying all correct answer options based on the reference text and full content provided. 
Example: If the correct answer is A and B, the correct_answers should be ["A", "B"].
If there is no correct answer, the correct_answers should be an empty list [].

Question: {question_data['question']}

Option A: {question_data['optionA']}
Option B: {question_data['optionB']}
Option C: {question_data['optionC']}
Option D: {question_data['optionD']}
Option E: {question_data['optionE']}
"""
\end{lstlisting}

\begin{lstlisting}[language=json,firstnumber=1,label={lst:classifier_prompt},caption={Industry classification prompt}]
f"""
You are a sustainability reporting expert specializing in corporate sustainability reports using IFRS standards. Your task is to identify industries directly related to a given question, referring to these industry descriptions:

{INDUSTRY_DESCRIPTIONS_STRING}

Guidelines:
1. Only return industries that are DIRECTLY and PRIMARILY relevant to the company's core business activities mentioned in the question.
3. Be extremely precise: do not include industries that are only tangentially related or those that the company might interact with but are not part of its primary operations.
4. Consider the context of corporate sustainability reporting when making your decision.
5. If multiple industries are relevant, limit your selection to the 1-3 most applicable ones.
6. Avoid including industries that might be part of the supply chain or waste management unless they are explicitly stated as a core part of the company's operations.
GIVE at least 1 undustry and at most 3 industries. return the full code like b1-apparel-accessories-and-footwear.
"""
\end{lstlisting}

\end{appendices}

\bibliographystyle{unsrtnat}
\bibliography{refs}

\end{document}